\documentclass[journal]{IEEEtran} 
\ifCLASSINFOpdf
\else
\fi
\usepackage{svg}
\usepackage{pdfpages}
\usepackage{subcaption}
\usepackage{graphicx}
\usepackage{color}
\usepackage{cite}
\usepackage{comment}
\usepackage{caption}
\usepackage{booktabs}
\usepackage{url}
\usepackage{hyperref}

\usepackage{framed}
\usepackage{listings}
\usepackage{xcolor}
 
\definecolor{codegreen}{rgb}{0,0.6,0}
\definecolor{codegray}{rgb}{0.5,0.5,0.5}
\definecolor{codepurple}{rgb}{0.58,0,0.82}
\definecolor{backcolour}{rgb}{0.95,0.95,0.92}
 
\lstdefinestyle{mystyle}{
    backgroundcolor=\color{backcolour},   
    commentstyle=\color{codegreen},
    keywordstyle=\color{magenta},
    numberstyle=\tiny\color{codegray},
    stringstyle=\color{codepurple},
    basicstyle=\scriptsize,
    breakatwhitespace=false,         
    breaklines=true,                 
    captionpos=b,                    
    keepspaces=true,                 
    numbers=left,                    
    numbersep=5pt,                  
    showspaces=false,                
    showstringspaces=false,
    showtabs=false,                  
    tabsize=2
}
 
\usepackage{ulem}

\begin{document}
%
\title{There Is No Data Like More Data -- Datasets for Deep Learning in Earth Observation}
%
%
%

\author{Michael Schmitt\textsuperscript{1}, Seyed Ali Ahmadi\textsuperscript{2}, Yonghao Xu\textsuperscript{3}, G\"ulsen Taskin\textsuperscript{4}, Ujjwal Verma\textsuperscript{5}, Francescopaolo Sica\textsuperscript{1}, Ronny H\"ansch\textsuperscript{6}\\\vspace{0.3pt}
\textsuperscript{1}Department of Aerospace Engineering, University of the Bundeswehr Munich, Neubiberg, Germany\\
\textsuperscript{2}Faculty of Geodesy and Geomatics Engineering, K. N. Toosi University of Technology, Tehran, Iran\\
\textsuperscript{3}Institute of Advanced Research in Artificial Intelligence (IARAI), Vienna, Austria\\
\textsuperscript{4}Institute of Disaster Management, Istanbul Technical University, Istanbul, Turkey\\
\textsuperscript{5}Department of Electronics and Communication Engineering, Manipal Institute of Technology Bengaluru, Manipal Academy of Higher Education, Manipal, India\\
\textsuperscript{6}Department of SAR Technology, Microwave and Radar Institute, German Aerospace Center (DLR), Wessling, Germany
}
\maketitle

\begin{abstract}
Carefully curated and annotated datasets are the foundation of machine learning, with particularly data-hungry deep neural networks forming the core of what is often called Artificial Intelligence (AI). Due to the massive success of deep learning applied to Earth Observation (EO) problems, the focus of the community has been largely on the development of ever-more sophisticated deep neural network architectures and training strategies largely ignoring the overall importance of datasets. For that purpose, numerous task-specific datasets have been created that were largely ignored by previously published review articles on AI for Earth observation. With this article, we want to change the perspective and put machine learning datasets dedicated to Earth observation data and applications into the spotlight. Based on a review of the historical developments, currently available resources are described and a perspective for future developments is formed. We hope to contribute to an understanding that the nature of our data is what distinguishes the Earth observation community from many other communities that apply deep learning techniques to image data, and that a detailed understanding of EO data peculiarities is among the core competencies of our discipline.
\end{abstract}


%
\IEEEpeerreviewmaketitle

\section{Introduction}

Deep learning techniques have enabled drastic improvements in many scientific fields, especially in those dedicated to the analysis of image data, e.g. computer vision or remote sensing. 
While it was possible to train \textit{shallow} learning approaches on comparably small datasets, \textit{deep} learning requires large-scale datasets to reach the desired accuracy and generalization performance. 
Therefore, the availability of annotated datasets has become a dominating factor for many cases of modern Earth observation data analysis that develops and evaluates powerful, deep learning-based techniques for the automated interpretation of remote sensing data. 

The main goal of general computer vision is the analysis of optical images, such as photos, which contain everyday objects, e.g., furniture, animals, or road signs. Remote sensing involves a larger variety of sensor modalities and image analysis tasks than conventional computer vision, rendering the annotation of remote sensing data more difficult and costly. 
Besides classical optical images, multi- or hyperspectral sensors and different kinds of infrared sensors, active sensor technologies such as laser scanning, microwave altimeters, and synthetic aperture radar (SAR) are regularly used, too. The fields of application range from computer vision-like tasks, such as object detection and classification, to semantic segmentation (mainly for land cover mapping) to specialized regression tasks grounded in the physics of the used remote sensing system. To provide an illustrative example, a dataset for biomass regression from interferometric SAR data will adopt imagery and annotations very different from the ones needed for the semantic segmentation of urban land cover types from multispectral optical data. 
Thus, while extensive image databases, such as ImageNet\footnote{As a prime example for an annotated computer vision dataset, ImageNet contains more than 14 million images depicting objects from more than 20,000 categories.}, have been created already more than ten years ago and form the backbone of many modern machine learning developments in computer vision, there is still no similar dataset or backbone network in remote sensing. 
This lack of generality renders the generation of an ImageNet-like general EO dataset extremely complicated and thus costly: Instead of photographs openly accessible in the internet, many different -- and sometimes quite expensive -- sensor data would have to be acquired, and instead of \textit{mechanical turks} (see box) trained EO experts would have to be hired to link these different sensor data to the multitude of different, domain- and task-specific annotations. Therefore, until now the trend in machine learning applied to EO data is characterized by the generation of numerous remote sensing datasets, each consisting of a particular combination of sensor modalities, applications, and geographic locations. Yet, a review of these developments is still missing in the literature. The only papers that make a small step towards a general review of benchmark datasets are \cite{Cheng2017,Hong2021,Li2020,Long2021}. All of them provide some sort of review, but always limited to a very narrow aspect, e.g. object detection or scene classification. Furthermore, their focus is on machine learning approaches and their corresponding datasets, while the historical evolution of datasets is neither discussed in detail, nor from a sensor- and task-agnostic point-of-view.

\noindent\fbox{%
    \parbox{\linewidth}{%
    \textbf{The Mechanical Turk}\\
       The name \textit{mechanical turk} comes from a fraudulent chess-playing machine developed in the 18th century. Chess-players were made to believe they played against the machine, but were in fact competing against a person hidden inside it. Today, the term mostly refers to Amazon Mechanical Turk (MTurk), a crowdsourcing website run by the corporation Amazon. On MTurk, users can hire remotely located crowd-workers to perform desired tasks. MTurk is frequently used to create manual annotations for supervised machine learning tasks.   
    }}%

As an extension of our IGARSS 2021 contribution \cite{Schmitt2021}, this paper intends to close this gap by

\begin{itemize}
    \item reviewing current developments in the creation of datasets for deep learning applications in remote sensing and Earth observation,
    \item structuring existing datasets and discussing their properties,
    \item providing a perspective on future requirements.
\end{itemize}

In this context, we additionally present the Earth Observation Database (EOD) \cite{schmitt2022eod}, which is the result of the effort and cooperation of voluntary scientists within the Image Analysis and Data Fusion (IADF) Technical Committee of IEEE GRSS. This database aims to function as a centralized tool that organizes the meta information about existing datasets in a community-driven manner.

\section{Evolution of EO-oriented Machine Learning Datasets}
\label{sec:evol}

\subsection{Historical Development}

\begin{figure}[t]
    \centering
    \includegraphics[width=\linewidth]{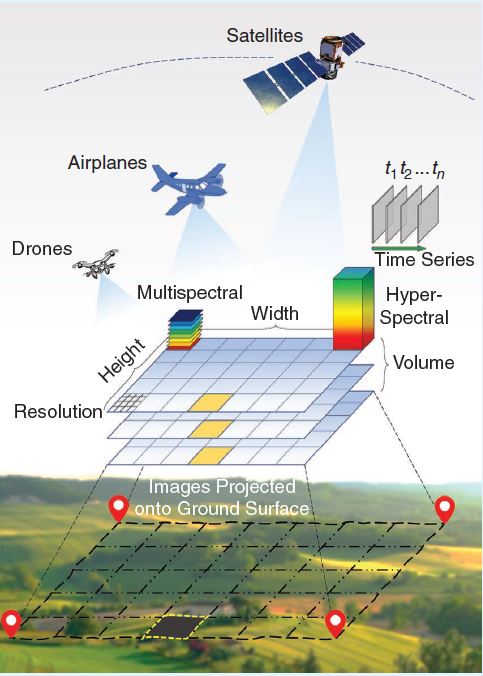}
    \caption{Schematic illustration of the proposed size measure used to characterize datasets i.e. pixels are only counted once in the spatial coverage provided by a dataset. For a more detailed definition, see the corresponding box.}
    \label{fig:sizemeasure}
\end{figure}

High-quality benchmark datasets have played an increasingly important role in Earth observation for quite some time and are one of the driving factors for the recent success of deep learning approaches to analyze remote sensing data. 
As such, they can be seen as a tool complementary to methodological advancements to push accuracy, robustness, and generalizability. 
This section reviews and summarizes the historical development of EO-oriented Machine Learning (ML) datasets to provide insights into the evolution of this "tool", ranging from its historical beginnings to the current state of the art. 

The beginnings of ML applied to remote sensing focused on specific applications. Datasets were mainly built by considering a very localized study site, a few specific sensor modalities, and a relatively small number of acquired samples. Therefore, the first datasets are relatively small compared to what is now considered a benchmarking dataset. Training, validation, and testing samples were often taken from the same image. 
Even with the use of sophisticated shallow learning models, but especially since the advent of deep learning, such small datasets were no longer sufficient for proper training and evaluation. The need for extended datasets led to the creation of larger datasets containing multiple images, often acquired at different geographic locations. 

\noindent\fbox{%
    \parbox{\linewidth}{%
    \textbf{How to Measure the Size of a Dataset}\\
       In this paper, we look at the \textit{size} of datasets from two perspectives:
\begin{itemize}
    \item \textit{Size}: Data volume in terms of the number of spatial pixels. We count the number of pixels in the highest available image resolution while ignoring multi-band, multi-channel, and multi-sensor data. In other words, pixels are only counted once in the spatial coverage provided by the dataset.
    \item \textit{Volume}: Data volume in terms of storage. The amount of disk space required for a dataset is a proxy for image resolution and provided modalities (e.g., multiple bands and sensor types).
\end{itemize}
    Fig.~\ref{fig:sizemeasure} highlights the different factors that affect the \textit{volume} and \textit{size} of a dataset: The number of bits per pixel (radiometric resolution), the number of spectral bands (spectral resolution, i.e., RGB, multispectral, or hyperspectral), the number of images during a specific time period (temporal resolution), and the number of pixels per unit area (spatial resolution). As mentioned above, the \textit{size} is directly related to the unique number of ground projected resolution cells. A larger dataset in terms of \textit{size} corresponds to images with higher resolutions or broader coverage. 
}}%

\begin{figure*}
    \centering
   \includegraphics[width=0.9\linewidth]{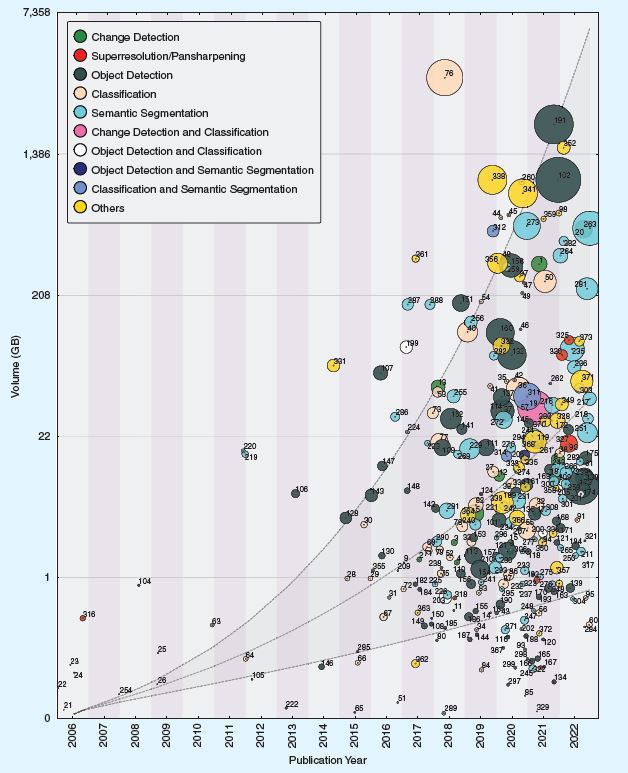}
    \caption{Distribution of remote sensing datasets over the years. The x-axis shows the publication year (the datasets are placed within the region of their publication year with a small random offset to minimize the visual overlap in the graph), while the y-axis represents the volume of each dataset in gigabytes on a logarithmic scale. The circle radius indicates the dataset size in terms of the number of pixels. Colors denote the type of task addressed by a dataset. Each circle is accompanied by an index, allowing to identify the dataset in the database (see Tab.~\ref{tab:database}) that provides further information.}
    \label{fig:evolution}
\end{figure*}

Fig.~\ref{fig:evolution} illustrates the evolution of benchmark datasets for ML in EO by showing this temporal development. 

To provide a broad view about the recent evolution of benchmarking datasets, we gathered an extensive list of datasets available in the EO community, resulting in a large collection with currently 400 (i.e., 380 image-based and 20 point-cloud-based datasets) entries (see Tab.~\ref{tab:database}), including related metadata. 
We point out that, while being extensive, this list is far from being complete due to the fact that a large amount of new datasets are published every year. 
Furthermore, the metadata required to generate the plot of Fig.~\ref{fig:evolution} are available only for a subset of 290 datasets (roughly 73\%).  
In the horizontal axis we indicate the year of publication.
The vertical axis shows the \emph{volume} of a dataset, while the circle radius reflects the number of \emph{spatial pixels} covered by a dataset. For a more detailed explanation of how we measure dataset \emph{size}, please refer to Fig.~\ref{fig:sizemeasure} and the corresponding box. 
Fig.~\ref{fig:evolution} provides a straightforward overview of the proportion between \emph{size}, spatial dimension and, therefore, about the overall information content given by features such as resolution, sensor modalities, number of bands/channels, etc. 
Each circle is accompanied by an index, allowing to identify the dataset in the database (see Tab.~\ref{tab:database}) that provides further information. As there are a few datasets published before 2006, the first column denotes the time of 2006 and earlier (denoted as $<$2006). 
Note that we use the category "Others" for datasets that do not belong to any of the other categories and are too rare to form their own category. Examples for the "Others" category are datasets on cloud removal visual question answering and parameter estimation tasks such as hurricane wind speed prediction, satellite pose estimation, and vegetation phenological change monitoring.
The dashed line illustrates an exponential growth of benchmark datasets created by and for the EO community. 

This map on the evolution of remote sensing datasets offers several interesting insights: 
\begin{enumerate}
    \item \textbf{The Beginnings}\\
    Additionally to the first IEEE GRSS Data Fusion Contest in 2006 (Tab.~\ref{tab:database}-\#316), there are a few other pioneering datasets that have fostered ML research applied to remote sensing data in its early stages, e.g.
\begin{itemize}
    \item Hyperspectral datasets (Indian Pines, Salinas Valley, and Kennedy Space Center,  (Tab.~\ref{tab:database}-\#21, \#22, and \#24)): Published before 2005, these datasets triggered the ML era in remote sensing. Covering a very small area on the ground and having a very small number of pixels, such datasets are not suitable for training DL models (or have to be used with excessive caution). On the other hand, due to their rich hyperspectral information, they are still being used for tasks such as dimensionality reduction and feature extraction.
    \item UC Merced dataset (Tab.~\ref{tab:database}-\#63) \cite{yang2010bag}: Published in 2010, it is the first dataset dedicated to scene classification. 
    \item ISPRS Potsdam/Vaihingen dataset (Tab.~\ref{tab:database}-\#219/220) \cite{rottensteiner2012isprs}: Published in 2012, it was initially intended to benchmark semantic segmentation approaches tailored to aerial imagery. Later it has also been used for other tasks, e.g., single-image height reconstruction (e.g. in \cite{Ghamisi2018}).
    \item SZTAKI-AirChange dataset (Tab.~\ref{tab:database}-\#105) \cite{benedek2011building}: Published in 2011, it is one of the earliest datasets designed for object detection. 
\end{itemize}
 All of those pioneering datasets have seen massive use in the early machine-learning-oriented Earth observation literature. It is interesting to note that pan-sharpening, scene classification, semantic segmentation, and object detection were the first topics in remote sensing to be addressed with machine learning based methodologies.

 \item \textbf{The Deep Learning Boom}
 As discussed by several review articles on deep learning and artificial intelligence applied to Earth observation \cite{Zhang2016,Zhu2017,Yuan2020}, the year 2015 marked the beginning of the deep learning boom in the EO community. This is well reflected by a significantly rising number of datasets published from that year onwards. It is furthermore confirmed by the fact that the dataset sizes both in terms of \emph{spatial pixels} and \emph{data volume} started to increase significantly from about that time.

\item \textbf{The Diversity of Tasks}
From the early days on, machine learning-oriented Earth observation datasets were designed for a multitude of different tasks. The historical evolution depicted in Fig.~\ref{fig:evolution} further shows that object detection and semantic segmentation are the most popular tasks, with a significant increase of datasets dedicated to minority categories (denoted by \emph{"Others"}) from about 2019 on. This indicates that the rise of deep learning in Earth observation broadens the overall application scope of the discipline. 
\end{enumerate}

\begin{figure}
    \begin{subfigure}[b]{0.45\textwidth}
        \centering
        \includegraphics[width=\linewidth]{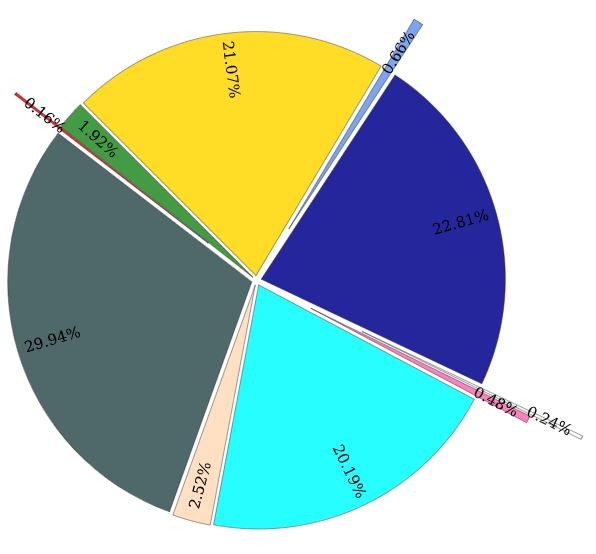}
        \caption{\textit{Volume} among different remote sensing tasks. With 30\%, Object Detection is the predominant task followed by Semantic Segmentation.} 
        \label{fig:PieChart-a}
    \end{subfigure}
    \hfill
    \begin{subfigure}[b]{0.45\textwidth}
        \centering
        \includegraphics[width=\linewidth]{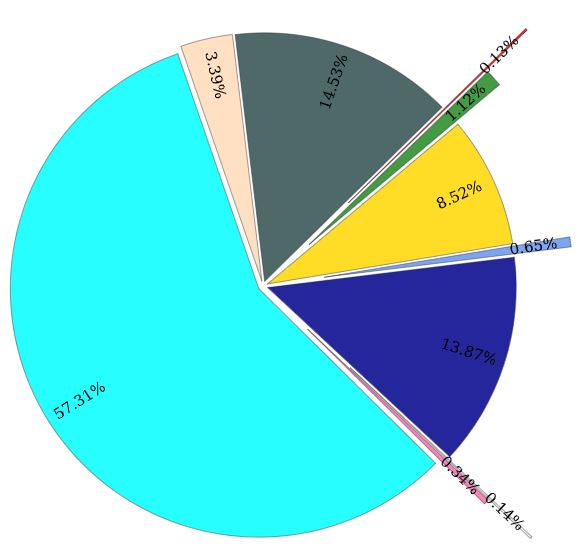}
        \caption{
        \textit{Size} among different remote sensing tasks. In contrast to Fig.~\ref{fig:PieChart-a}, Semantic Segmentation is the prevailing task illustrating that corresponding datasets involve more complex scenarios such as leveraging multiple sensors or spectral bands.}
        \label{fig:PieChart-b}
    \end{subfigure}
    \begin{subfigure}[b]{0.45\textwidth}
        \centering
        \includegraphics[width=\linewidth]{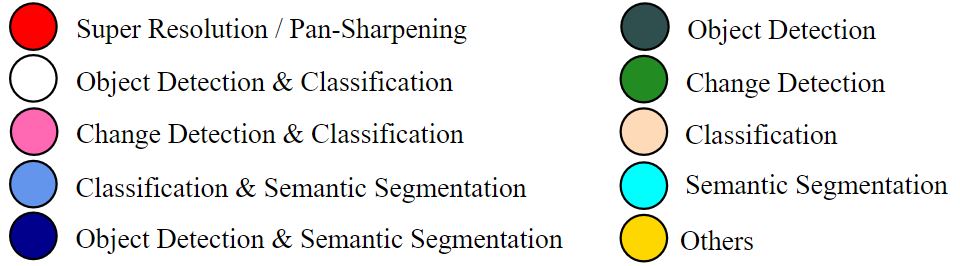}
    \end{subfigure}
  \caption{Distribution of EO dataset sizes over typical remote sensing tasks, expressed in (a)\textit{volume} and (b) \textit{size} as defined in Fig.~\ref{fig:sizemeasure} and the corresponding box. Object Detection and Semantic Segmentation are the dominating image analysis tasks in ML-centered EO.}
\label{fig:distribution}
\end{figure}

Fig.~\ref{fig:distribution} provides further insights into the size distribution of available datasets in terms of the (1) number of pixels and (2) data volume in GB for typical remote sensing tasks. Illustrating these two different aspects allows a further understanding of the nature of the data. For example, datasets counting a similar number of spatial pixels may differ in data volume, which can, for example, indicate the use of multimodal imagery.
Object detection offers the largest data volume among the existing benchmarking datasets, which again confirms its popularity in deep learning-oriented remote sensing research. However, in terms of number of pixels, semantic segmentation takes the lead, indicating that a larger spatial coverage is usually involved for this type of application.

\begin{figure*}
    \centering
    \includegraphics[width=\linewidth]{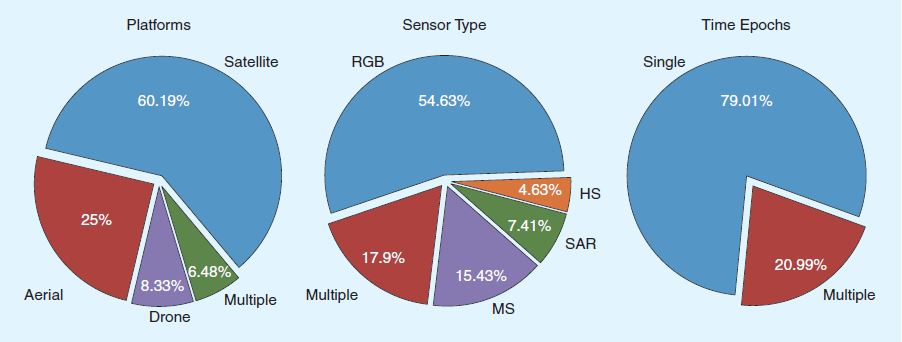}
    \caption{Distribution of available EO datasets over different platforms, sensor types, and number of acquisition times. Single-image RGB images acquired by satellites are clearly the dominating modality.}
    \label{fig:pie_3}
\end{figure*}

\subsection{Platforms}
\begin{figure*}
    \centering
    \includegraphics[width=0.6\textwidth]{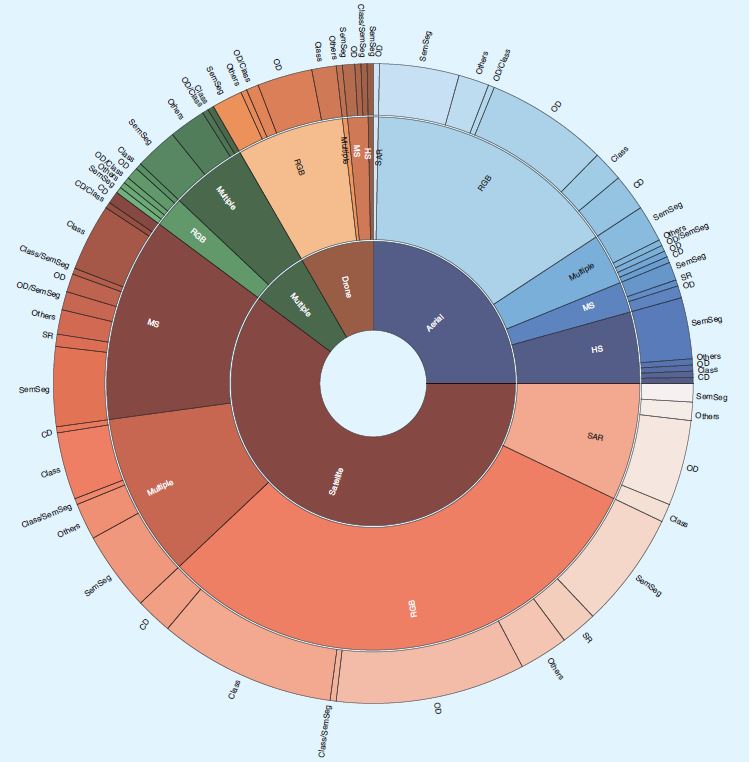}
    \caption{Distribution of tasks between sensors and platforms. Platforms are in the inner ring, sensors are distributed in the middle ring, and the outer ring shows different tasks per sensor.}
    \label{fig:nested}
\end{figure*}

A direct overview of the occurrence of a type of platform and sensor is given in Fig.~\ref{fig:pie_3}. Satellite platforms are the most common, followed by airborne platforms and drones. Optical data cover more than half of the datasets, while all other sensors are almost equally distributed. Interestingly, $20\%$ of the datasets provide time series, while the rest are single-temporal acquisitions.  
Complementing this, Fig.~\ref{fig:nested} highlights the distribution of tasks between sensors and platforms. The inner ring indicates the platform type, which then splits into different sensor types in the middle ring, and finally denotes the targeted tasks in the outer ring, respectively. This graph shows that the datasets acquired by UAVs and aircrafts are mainly dedicated to optical sensors, while satellite-based Earth observation has a much wider and more homogeneous distribution across all sensor and application types.

\begin{figure*}
    \centering
    \includegraphics[width=0.6\linewidth]{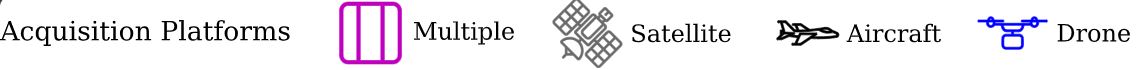}
    \begin{subfigure}[b]{\textwidth}
    \centering
    \includegraphics[width=0.9\linewidth]{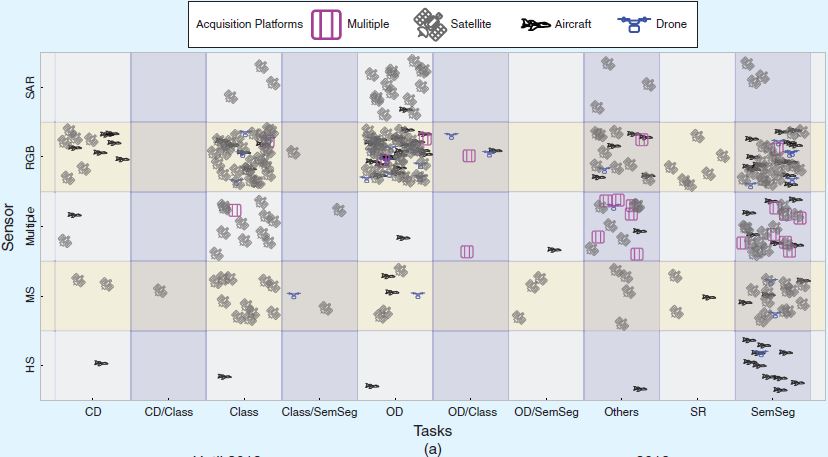}
    \caption{Combinations of different platforms, sensors, and tasks accumulated over the years.}
    \label{fig:platforms_all}
    \end{subfigure}
    \begin{subfigure}[b]{\textwidth}
    \centering
    \includegraphics[width=0.9\linewidth]{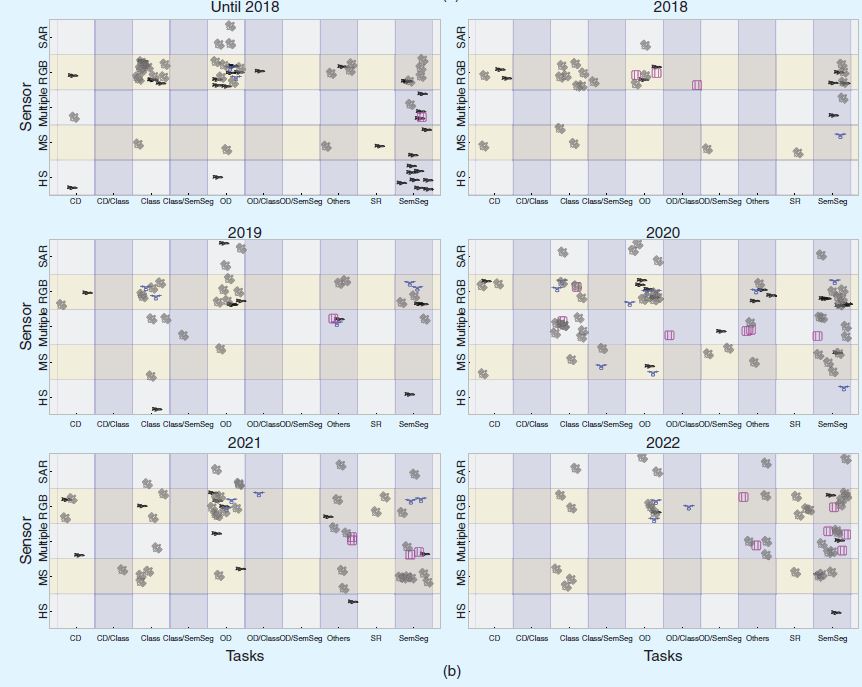}
    \caption{Combinations of different platforms, sensors, and tasks for different years.}
    \label{fig:platforms_annual}
    \end{subfigure}
    \caption{Different tasks (Change Detection (CD), Classification (Class), Semantic Segmentation (SemSeg), Object Detection (OD), Super Resolution and Pan-sharpening (SR)) and task combinations (denoted by '/') make use of very different platforms and sensors.}
    \label{fig:platforms}
\end{figure*}
 
Fig.~\ref{fig:platforms_all} and Fig.~\ref{fig:platforms_annual} further specify the previous findings by showing how the datasets acquired by different platforms are distributed across both tasks and sensors. 
While the former provides an overview, the latter adds a temporal aspect by showing the corresponding applications per year. The x- and y-axes represent tasks and sensors, respectively, while the different markers indicate the type of acquisition platforms. From these plots we can affirm that object detection and classification tasks are mainly performed on optical images. At the same time, semantic segmentation is fairly evenly distributed between optical, multispectral, and other sensor combinations.
SAR images are mainly acquired from satellite platforms, while hyperspectral datasets are almost always acquired from airborne systems. UAVs mainly carry optical sensors in the context of semantic segmentation.
Some tasks, such as super-resolution, naturally make use of multimodal data, e.g., optical and hyperspectral imagery.  
The year-by-year graph in Fig.~\ref{fig:platforms_annual} shows that super-resolution datasets, together with UAV-based acquisitions, have received more attention in recent years. On the other hand, the EO community has not seen many new hyperspectral datasets since 2018. Optical sensors were the most common source of information, while after 2020 an increasing number of datasets were also acquired from other sensors, such as SAR or hyperspectral systems. 

Figures~\ref{fig:pie_3},~\ref{fig:nested}, and~\ref{fig:platforms_all} show that the number of datasets acquired from "multiple" platforms or sensors is still the minority which provides evidence for the earlier statement that state-of-the-art datasets are usually designed to respond to a specific task in EO applications.
These figures also show which combination of EO tasks, platforms, and sensors is currently underrepresented. In particular Fig.~\ref{fig:platforms_all} shows three main gaps: (1) SAR change detection, (2) SAR super-resolution, and (3) hyperspectral super-resolution. From a sensor perspective alone, the lack of airborne SAR datasets and drone-based hyperspectral benchmarks are other obvious gaps. 

\begin{figure*}
    \centering
    \includegraphics[width=\linewidth]{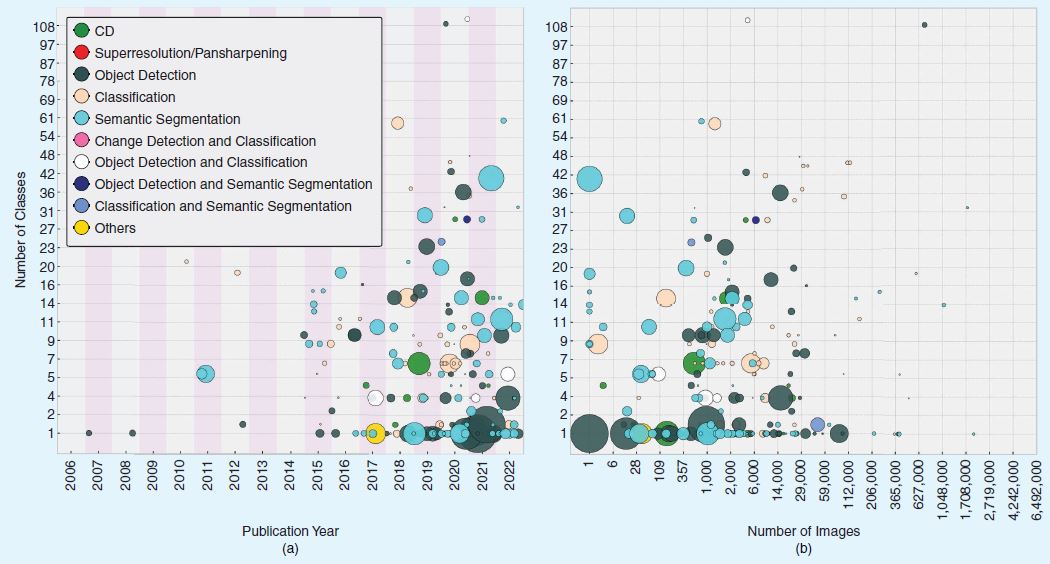}
    \caption{The number of classes provided by the reference data of a given dataset does not only vary for different tasks (e.g. Object Detection is dominated by datasets with only a single class), but also with publication year (left) and the number of images (right) of a dataset.    
    }
    \label{fig:subplots}
\end{figure*}

\subsection{Geographic Diversity}
\begin{figure*}
    \centering
    \includegraphics[width=\textwidth]{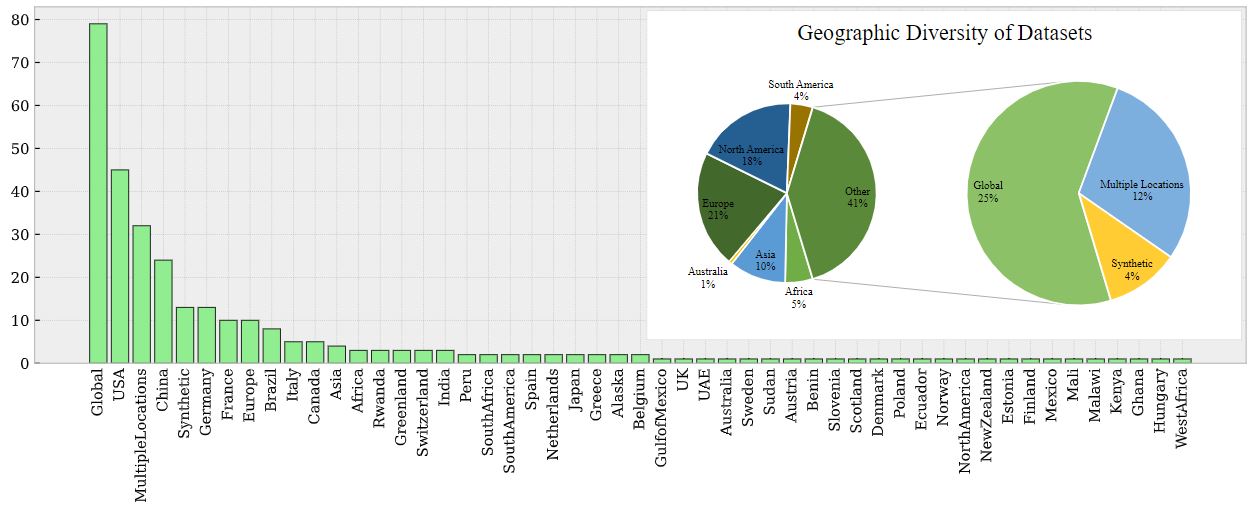}
    \caption{Geographic distribution of EO benchmark datasets (that provided clear location information).}
    \label{fig:diversityBar}
\end{figure*}

The geographic diversity and spatial distribution of the Earth observation imagery is an important attribute of a benchmark dataset as it is directly related to the geographic generalization performance of data-driven models. 
Fig.~\ref{fig:diversityBar} shows the geographic distribution of datasets (i.e. of roughly $\sim$300 datasets (75\%) from Tab.~\ref{tab:database} that provided (or where we could find) their geographic information). 
Many of the datasets are globally distributed (\textit{Global}) and contain images from around the world, while others cover a limited set of multiple cities or countries (\textit{Multiple Locations}). 
Maybe surprisingly, \textit{synthetic} datasets show a dominating presence as well, illustrating the benefits of being able to control image acquisition parameters (such as viewing angle), environmental factors (such as atmospheric conditions, illumination, and cloud cover), and scene content (e.g. type, size, shape, and number of objects). 
Fig~\ref{fig:diversityBar} illustrates an important and within the EO community seldom discussed issue: There exists a strong geographic bias within the available EO datasets. While 25\% of the datasets contain samples from globally distributed locations, nearly 40\% of available datasets are covering regions in Europe (21\%) and North America (18\%) only. Asia is still covered by 10\%, however, Africa (5\%), South America (4\%), and Australia (1\%) are barely included. This raises the question whether many of the findings and conclusions in corresponding research articles would generalize to these geographic areas. In any case, the need for more spatially diverse datasets becomes apparent, in particular covering also underdeveloped countries. 

\subsection{Source of the Reference Data}
Another important aspect is the source of the reference data. 
While most scientific papers introducing a new benchmark dataset are very detailed regarding source and properties of the provided image data, they often contain only sparse information about the exact process of how the reference data was obtained. 
However, knowledge about the source of the reference data as well as about measures for quality assessment and quality control are essential to judge the potential of the benchmark and how to interpret obtained results. 
For example, achieving a high "accuracy" on a dataset that contains a high amount of annotation errors in its reference data only means that the model is reproducing the same bias as the method used to annotate the data but is not actually producing accurate results. 
Furthermore, information about the source of the reference data is not only scarce but also very heterogeneous. 
Examples include manual annotation for semantic segmentation (e.g. OpenEarthMap (Tab-\ref{tab:database}-\#253)) or object detection (e.g. BIRDSAI (Tab-\ref{tab:database}-\#127), DroneCrowd (Tab-\ref{tab:database}-\#138), the use of keywords in image search engines for classification (e.g. AIDER (Tab-\ref{tab:database}-\#34)), leveraging existing resources (e.g. BigEarthNet (Tab-\ref{tab:database}-\#40) uses the CORINE LULC map, MapInWild (Tab-\ref{tab:database}-\#264) uses the World Database of Protected Areas, AI-TOD (Tab-\ref{tab:database}-\#162) uses other existing datasets, ship detection datasets usually use AIS data, OpenSentinelMap (Tab-\ref{tab:database}-\#282) uses OSM), automatic reference methods (e.g. HSI Oil Spill (Tab-\ref{tab:database}-\#211) and Sen12MS-CR-TS (Tab-\ref{tab:database}-\#98), or a mixture of these (e.g. RSOC (Tab-\ref{tab:database}-\#117) uses existing datasets (DOTA, Tab-\ref{tab:database}-\#109) as well as manual annotations for new images).
The quality of class definitions varies among different datasets, often depending on whether a dataset is designed for a specific real-world application (in which class definitions are driven by the application context) or whether it is just an example dataset to train and evaluate a machine-learning based approach (in which case class definitions are more arbitrary). 
If human interaction is involved (e.g. via manual annotations), annotation protocols or precise class definitions are often not shared (if they even exist). 
With the evolution of datasets, the quality of the meta-information about the reference data needs to increase together with the quality and quantity of the image data.

\subsection{Applications and Tasks}
Finally, we analyze a last aspect concerning the distribution of some characteristics of the datasets, namely the number of classes and images. In Fig.~\ref{fig:subplots}, we show on the left side the evolution of the number of classes per publication year. On the right side, we plot the number of classes against the number of images in the dataset. In both plots, the radius indicates the size of the image (width or height), while the color indicates the type of task. We can see that there is no clear trend or correlation between the year of publication and the number of classes in a dataset. Instead, we find that more recently published datasets increase the variety in the number of classes, again highlighting an increased interest in using benchmarking datasets in a wider range of applications.
Furthermore, although there is no clear correlation, we can confirm that increasing the number of classes reduces the size of the images and increases the number of images. More populated datasets, in terms of number of images, also have smaller image sizes and typically consider a smaller number of classes. Conversely, larger images are found in less populated datasets.

This overview is a snapshot of the currently published benchmarking datasets. Although the list of datasets is constantly growing, we believe that the observed trends will not change much over the next few years. Instead, we expect long-term developments that lead to two divergent aspects of datasets: specificity and generality. This will be further discussed in Section~\ref{sec:open}.

\begin{table*}
\begin{framed}
\begin{minipage}{\textwidth}
\vspace{0pt}
Point-cloud datasets are another large group of benchmark data that are widely used in the literature and industry. Within EO the most common source for point-cloud data are LiDAR (Light Detection and Ranging) sensors that use light in the form of laser pulses to measure the distance to the surface. The primary sources are Airborne, Terrestrial, and Mobile Laser Scanning devices (i.e., ALS, TLS, and MLS, respectively). 
Other sources of point clouds and 3D data include photogrammetric methods (Structure from Motion, Multi-View Stereo, dense matching approaches) and tomographic synthetic aperture radar (TomoSAR). 
Since 3D data typically comes with features very different to 2D image data, such datasets are beyond the scope of this paper. Nevertheless, Tab.~\ref{tab:lidarDB} provides a short list of example LiDAR/point-cloud datasets for interested readers.
\vspace{10pt}
\end{minipage}\\
\begin{minipage}{\textwidth}
\vspace{0pt}
    \tiny
    \centering
    \begin{tabular}{cccccccccc}
    \hline
        \textbf{Task} & \textbf{Platform} & \textbf{Time stamps} & \textbf{Name} & \textbf{Pub. Date} & \textbf{Point Density (pts/m2)} & \textbf{\#Classes} & \textbf{\#Points} & \textbf{Volume (MB)} \\ \hline
        \textbf{Change Detection} & ALS & Multiple & Abenberg ALS & 2013 & 16 & - & 5400000 & 258 \\ 
        \textbf{Classification} & ALS & Single & NEWFOR & 2015 & varies & 4 & - & 97 \\ 
        \textbf{Classification} & ALS & Single & DFC19 & 2019 & - & 6 & 167400000 & 613 \\ 
        \textbf{Classification} & ALS & Single & ISPRS 3D Vaihingen & 2014 & 8 & 9 & 780879 \\ 
        \textbf{Classification} & Multiple & Single & ArCH & 2020 & varies & 10 & 136138423 & - \\ 
        \textbf{Classification / Semantic Segmentation} & ALS & Single & DublinCity & 2019 & 240 to 348 & 13 & 260000000 & 3000 \\ 
        \textbf{Filtering} & ALS & Single & OpenGF & 2021 & 6 and 14 & 3 & 542100000 & 2280 \\ 
        \textbf{Object Detection /Semantic Segmentation} & TLS & Single & LiSurveying & 2021 & varies & 54 & 2450000000 & - \\ 
        \textbf{Others} & ALS & Single & RoofN3D & 2018 & 4.72 & 3 & 118100 & -  \\ 
        \textbf{Semantic Segmentation} & ALS & Single & LASDU & 2020 & 3-4 & 6 & 3120000 & - \\ 
        \textbf{Semantic Segmentation} & ALS & Single & DALES & 2020 & 50 & 8 & 505000000 & 4000 \\ 
        \textbf{Semantic Segmentation} & ALS & Single & DALES Object & 2021 & 50 & 8 & 492000000 & 5000 \\ 
        \textbf{Semantic Segmentation} & Drone & Single & Campus3D & 2020 & varies & 24 & 937000000 & 2500 \\ 
        \textbf{Semantic Segmentation} & Drone & Multiple & Hessigheim 3D & 2021 & 800 & 11 & 73909354 & 5950 \\ 
        \textbf{Semantic Segmentation} & Drone & Single & WildForest3D & 2022 & 60 & 6 & 7000000 & 81 \\ 
        \textbf{Semantic Segmentation} & MLS & Single & Toronto3D & 2020 & 1000 & 8 & 78300000 & 1100 \\ 
        \textbf{Semantic Segmentation} & MLS & Multiple & HelixNet & 2022 & - & 9 & 8850000000 & 235700 \\ 
        \textbf{Semantic Segmentation} & Photogrammetry & Single & SensatUrban & 2020 & - & 13 & 2847000000 & 36000 \\ 
        \textbf{Semantic Segmentation} & Photogrammetry & Single & STPLS3D & 2022 & - & 20 & - & 36600 (images:700000) \\ 
        \textbf{Semantic Segmentation} & TLS & Single & Semantic 3D & 2017 & - & 8 & 4000000000 & 23940 \\ \hline
    \end{tabular}
    \caption{
    While a thorough analysis of LiDAR datasets is beyond the scope of this survey, we do provide an overview of several example datasets.
    }
    \label{tab:lidarDB}
\end{minipage}
\end{framed}
\end{table*}

\begin{table*}[!ht]
    \centering
    \tiny
    \begin{tabular}{ccccccccccccc}
    \hline
        \textbf{Index} & \textbf{Task} & \textbf{Platform} & \textbf{Sensor Type} & \textbf{Name} & \textbf{Pub. Date} & \textbf{Timestamps} & \textbf{\#Images} & \textbf{Img. Size} & \textbf{Size} & \textbf{\#Classes} & \textbf{Volume (MB)} \\ \hline
        \textbf{1} & Change Detection & Aerial & Multiple & DFC21-MSD & 2021 & Multiple & 2250 & 4000 & 36000000000 & 15 & 325000 \\ 
        \textbf{10} & Change Detection & Satellite & Multiple & DFC09 & 2009 & Multiple & 2 & 98 & ~ & ~ & ~ \\ 
        \textbf{11} & Change Detection & Satellite & Multispectral & OneraCD & 2018 & Multiple & 24 & 600 & 8640000 & 2 & 489 \\ 
        \textbf{15} & Change Detection & Satellite & Optical & LEVIR-CD & 2020 & Single & 637 & 1024 & 667942912 & 1 & 2700 \\ 
        \textbf{21} & Classification & Aerial & Hyperspectral & Indian Pines & 2000 & Single & 1 & 145 & 21025 & 16 & 6 \\ 
        \textbf{22} & Classification & Aerial & Hyperspectral & Salinas & 2000 & Single & 1 & 365 & 111104 & 16 & 27 \\ 
        \textbf{24} & Classification & Aerial & Hyperspectral & Kennedy Space Center & 2005 & Single & 1 & 550 & 311100 & 13 & 57 \\ 
        \textbf{34} & Classification & Drone & Optical & AIDER & 2019 & Single & 2645 & 240 & 152352000 & 4 & 275 \\ 
        \textbf{40} & Classification & Satellite & Multiple & BigEarthNet-MM & 2019 & Single & 590326*12 & 120 & 1.02E+11 & 19 & 121000 \\ 
        \textbf{52} & Classification & Satellite & Multispectral & EuroSAT & 2018 & Single & 27000*13 & 64 & 1437696000 & 10 & 1920 \\ 
        \textbf{63} & Classification & Satellite & Optical & UC Merced & 2010 & Single & 2100 & 256 & 137625600 & 21 & 317 \\ 
        \textbf{69} & Classification & Satellite & Optical & AID & 2017 & Single & 10000 & 600 & 3600000000 & 30 & 2440 \\ 
        \textbf{76} & Classification & Satellite & Optical & FMoW & 2018 & Single & 523846 & - & 1.08E+12 & 63 & 3500000 \\ 
        \textbf{98} & Cloud Removal & Satellite & Multiple & SEN12MS-CR-TS & 2021 & Multiple & 53 & 4000 & 848000000 & - & 649000 \\ 
        \textbf{105} & Object Detection & Aerial & Optical & SZTAKI AirChange & 2012 & Multiple & 13 & 800 & 7920640 & 2 & 42 \\ 
        \textbf{109} & Object Detection & Aerial & Optical & DOTA v1.0 & 2018 & Single & 2806 & 4000 & 44896000000 & 15 & 18000 \\ 
        \textbf{114} & Object Detection & Aerial & Optical & DOTA v2.0 & 2020 & Single & 11268 & 4000 & 1.80E+11 & 18 & 34280 \\ 
        \textbf{117} & Object Detection & Aerial & Optical & RSOC & 2020 & Single & 3057 & 2500 & 3621481392 & 4 & - \\ 
        \textbf{127} & Object Detection & Drone & Multispectral & BIRDSAI & 2020 & Multiple & 162000 & 640 & 49766400000 & - & 43200 \\ 
        \textbf{138} & Object Detection & Drone & Optical & DroneCrowd & 2022 & Multiple & 33600 & 1920 & 69672960000 & 1 & 10400 \\ 
        \textbf{151} & Object Detection & Satellite & Optical & SpaceNet-4 & 2018 & Single & 60000 & 900 & 48600000000 & 186000 & ~ \\ 
        \textbf{162} & Object Detection & Satellite & Optical & AI-TOD & 2020 & Single & 28036 & - & - & - & 22000 \\ 
        \textbf{191} & Object Detection & Satellite & SAR & xView3-SAR & 2021 & Single & 991*2 & 27000 & 1.42E+12 & 2 & 2000000 \\ 
        \textbf{204} & Object Detection/Classification & Satellite & SAR & FUSAR-Ship & 2020 & Single & 5000 & 512 & 1310720000 & 113 & ~ \\ 
        \textbf{208} & Semantic Segmentation & Aerial & Hyperspectral & DFC08 & 2008 & Single & 5 & 194 & ~ & ~ & ~ \\ 
        \textbf{211} & Semantic Segmentation & Aerial & Hyperspectral & HOSD & 2022 & Single & 18*224 & 1700 & 3622355072 & 1 & 2200 \\ 
        \textbf{212} & Semantic Segmentation & Aerial & Multiple & DFC13 & 2013 & Single & 145 & 16 & 1000 & ~ & ~ \\ 
        \textbf{213} & Semantic Segmentation & Aerial & Multiple & DFC14 & 2014 & Single & 7 & ~ & ~ & ~ & ~ \\ 
        \textbf{214} & Semantic Segmentation & Aerial & Multiple & DFC15 & 2015 & Single & 7 & 10000 & 700000000 & ~ & ~ \\ 
        \textbf{215} & Semantic Segmentation & Aerial & Multiple & DFC18 & 2018 & Single & 1 & 20 & 19763 & ~ & ~ \\ 
        \textbf{217} & Semantic Segmentation & Aerial & Multiple & DFC22-SSL & 2022 & Single & 5000 & 2000 & 20000000000 & 14 & 42000 \\ 
        \textbf{219} & Semantic Segmentation & Aerial & Multispectral & ISPRS 2D - Potsdam & 2011 & Single & 38 & 6000 & 1368000000 & 6 & 16000 \\ 
        \textbf{220} & Semantic Segmentation & Aerial & Multispectral & ISPRS 2D - Vaihingen & 2011 & Single & 33 & 2200 & 156750000 & 6 & 17000 \\ 
        \textbf{246} & Semantic Segmentation & Multiple & Multiple & DFC17 & 2017 & Multiple & 17 & ~ & ~ & ~ & ~ \\ 
        \textbf{247} & Semantic Segmentation & Multiple & Multiple & SpaceNet-6 & 2020 & Single & 3401 & 900 & 2754810000 & 1 & 368 \\ 
        \textbf{253} & Semantic Segmentation & Multiple & Optical & OpenEarthMap & 2022 & Single & 5000 & 1024 & 5242880000 & 8 & 9100 \\ 
        \textbf{254} & Semantic Segmentation & Satellite & Multiple & DFC07 & 2007 & Single & 1 & 787 & 619369 & 19 & ~ \\ 
        \textbf{257} & Semantic Segmentation & Satellite & Multiple & DFC20 & 2020 & Single & 180662 & 17 & 19200 & ~ & ~ \\ 
        \textbf{261} & Semantic Segmentation & Satellite & Multiple & DFC21-DSE & 2021 & Single & 6000 & 16 & 1536000 & 4 & 18000 \\ 
        \textbf{264} & Semantic Segmentation & Satellite & Multiple & MapInWild & 2022 & Single & 1018*8 & 1920 & 30022041600 & 11 & 365000 \\ 
        \textbf{282} & Semantic Segmentation & Satellite & Multispectral & OpenSentinelMap & 2022 & Multiple & 137045 & 192 & 5052026880 & 15 & 445000 \\ 
        \textbf{286} & Semantic Segmentation & Satellite & Optical & SpaceNet-1 & 2016 & Single & 9735 & 650 & 4113037500 & 1 & 31000 \\ 
        \textbf{287} & Semantic Segmentation & Satellite & Optical & SpaceNet-2 & 2017 & Single & 24586 & 650 & 10387585000 & 1 & 182200 \\ 
        \textbf{288} & Semantic Segmentation & Satellite & Optical & SpaceNet-3 & 2017 & Single & 3711 & 1300 & 6271590000 & 1 & 182200 \\ 
        \textbf{292} & Semantic Segmentation & Satellite & Optical & SpaceNet-5 & 2019 & Single & 2369 & 1300 & 4003610000 & 1 & 84103 \\ 
        \textbf{294} & Semantic Segmentation & Satellite & Optical & SpaceNet-7 & 2020 & Multiple & 1525 & 1024 & 1599078400 & 1 & 20582 \\ 
        \textbf{301} & Semantic Segmentation & Satellite & Optical & SpaceNet-8 & 2022 & Multiple & 1200 & 1200 & 3456000000 & 2 & 6800 \\ 
        \textbf{312} & Semantic Segmentation, Classification & Satellite & Multiple & SEN12MS & 2019 & Single & 180662 & 256 & 11839864832 & - & 510000 \\ 
        \textbf{313} & Semantic Segmentation, Classification & Satellite & Multiple & DFC23 & 2023 & Single & 12 & ~ & ~ & ~ & ~ \\ 
        \textbf{316} & Super Resolution, Pan-Sharpening & Aerial & Multispectral & DFC06 & 2006 & Single & 6 & 5000 & 637500000 & - & 390 \\ 
        \textbf{318} & Super Resolution, Pan-Sharpening & Satellite & Multispectral & Proba-V Super Resolution & 2018 & Single & 1160 & 384 & 171048960 & - & 692 \\ 
        \textbf{319} & Super Resolution, Pan-Sharpening & Satellite & Multispectral & PAirMAX (Airbus) & 2021 & Single & - & - & - & - & 153 \\ 
        \textbf{324} & Super Resolution, Pan-Sharpening & Satellite & Optical & PAirMAX (MAXAR) & 2021 & Single & 14 & - & - & - & 386 \\ 
        \textbf{325} & Super Resolution, Pan-Sharpening & Satellite & Optical & WorldStrat & 2022 & Single & 3928 & 1054 & 4363678048 & - & 107000 \\ 
        \textbf{340} & Others & Multiple & Multiple & DFC19 & 2019 & Single & ~ & ~ & ~ & ~ & ~ \\ 
        \textbf{360} & Others & Satellite & Optical & DFC16 & 2016 & Multiple & ~ & ~ & ~ & ~ & ~ \\ \hline
    \end{tabular}
    \caption{
    While the complete collection of 380 datasets is too extensive to be included in the print version of this article, it can be found at \url{https://www.grss-ieee.org/wp-content/uploads/2023/05/EODatasets.pdf}. Here we include a part of this list that contains all datasets mentioned in the paper with their index in the complete database. 
    }
    \label{tab:database}
\end{table*}

\section{Exemplary Datasets}
\label{Sec:ExemDatasets}

This section provides short descriptions of a selection of Earth Observation datasets for various tasks: semantic segmentation, classification, object detection, change detection, and super-resolution/pan-sharpening. Given the vast amount of publicly available EO datasets, it is only possible to present some of them in this paper. Thus, this selection cannot be comprehensive and certainly follows a subjective view influenced by the experience of the authors. However, the selected datasets are representative for their respective tasks and were selected based on their observed relevance: They either are the largest in terms of \textit{size} (see Section~\ref{sec:evol} for a definition), the first to be introduced for a given task, or the most popular dataset for the chosen application. The popularity was determined based on the number of citations the original paper introducing the dataset received. 

The popularity of a dataset is influenced by multiple factors. 
One is certainly the size of a dataset, i.e. larger datasets are often preferred. However, there are exceptions. For instance, Functional Map of the World (FMoW) (Tab-\ref{tab:database}-\#76), introduced in 2018 \cite{christie2018functional}, is the largest dataset for remote sensing scene classification in terms of the number of images (1 million), but has yet to gain a high level of popularity (with 200 citations and 128 GitHub stars, compared to EuroSAT (Tab.~\ref{tab:database}-\#52) with 796 citations and 276 GitHub stars, or AID (Tab.~\ref{tab:database}-\#69) with 1000 citations, which are all published in 2018). Several other factors affect the popularity of a dataset, too, such as ease of access to the hosting servers (Google Drive, IEEE Dataport, Amazon AWS, University/Personal servers, etc.), accompanying documentation and presentation (standard metadata format, detailed description, availability of template code and support, suitable Search Engine Optimization, etc), or ease of downloading the data (temporary or permanent links, the bandwidth of hosting server, sign in/up requirements, etc.). Last but not least, an already established dataset is more likely to be used in new studies to enable comparison to related prior works, even if newer (and potentially better) datasets might exist.

Along with brief descriptions, this section provides insights into the different dataset characteristics.

\subsection{Semantic Segmentation}

\begin{figure*}
    \centering
    \includegraphics[width=\textwidth]{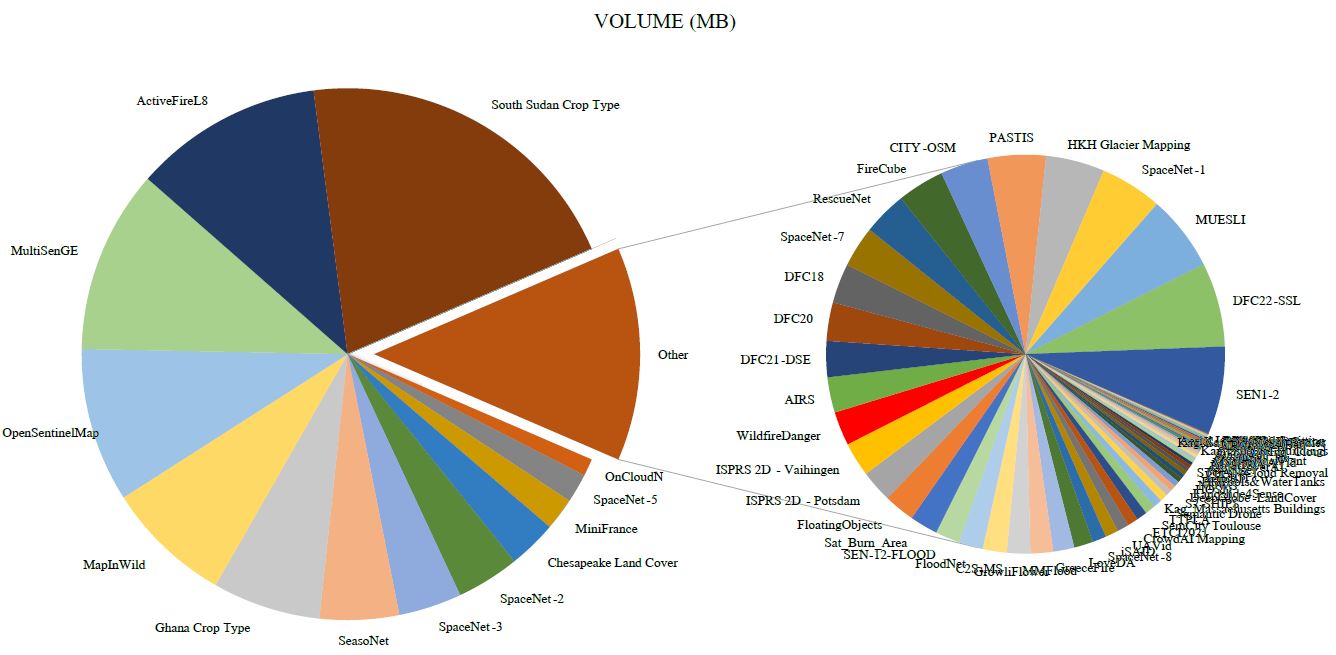}
    \caption{Relative volume distribution among datasets addressing semantic segmentation.}
    \label{fig:data_semseg}
\end{figure*}

Semantic segmentation refers to assigning a class label to each pixel in the image. Partitioning the image into semantically meaningful parts creates a more detailed map of the input image and provides a foundation for further analysis such as land use land cover (LULC) classification, building footprint extraction, landslide mapping, etc. \cite{Ioannis2021SegReview}. 
Fig.~\ref{fig:data_semseg} shows the relative volume of corresponding benchmark datasets and illustrates a wide spread of dataset sizes. Here, we present two examples: {\it SEN12MS} (focusing on medium-resolution satellite data and weak annotations) and the {\it ISPRS Vaihingen dataset} (focusing on very-high-resolution airborne data with high-quality annotations). 

\begin{figure}[ht]
  \centering
  \includegraphics[width=0.9\linewidth]{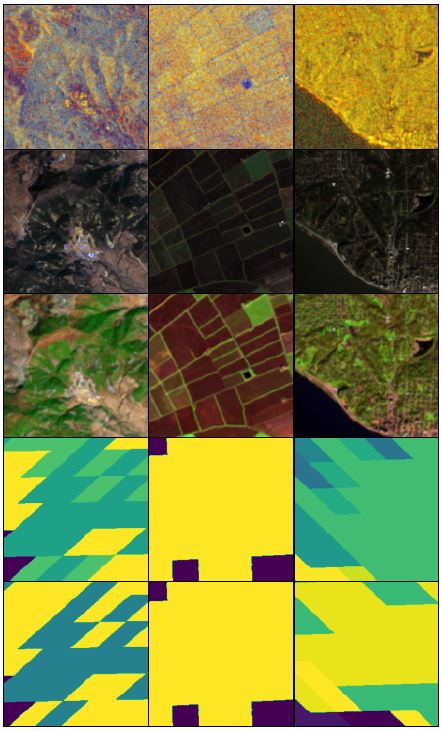}
  \caption{Several patch triplet examples of the SEM12MS dataset. Each row shows (from top to bottom): false color Sentinel-1 SAR (R: VV, G: VH, B: VV/VH), Sentinel-2 RGB, Sentinel-2 SWIR, IGBP Land cover, and LCCS Land cover.}
\label{fig:SEN12MS}
\end{figure}

{\it 1) SEN12MS}\footnote{\url{https://github.com/schmitt-muc/SEN12MS}} is among the most popular and largest datasets  (Tab.~\ref{tab:database}-\#312) in terms of volume of data as shown in Fig. \ref{fig:evolution}. A total of 541,986 image patches of size $256 \times 256$ pixels with high spectral information content  is present in this dataset \cite{Schmitt2019}. It contains dual-polarimetric SAR image patches from Sentinel-1, multispectral image patches from Sentinel-2, and  MODIS land cover maps (Fig. \ref{fig:SEN12MS}). The image patches are acquired at random locations around the globe and cover the four meteorological seasons of the northern hemisphere. The patches were further processed to ensure that the images did not contain any clouds, shadows, and artifacts. In addition to the images and the land cover maps, the results of two baseline CNN classifiers (ResNet50 and DenseNet121) are also discussed to demonstrate the usefulness of the dataset for land cover applications \cite{Sen12MSSchmitt2021}. 

\begin{figure}[ht]
  \centering
  \includegraphics[width=0.9\linewidth]{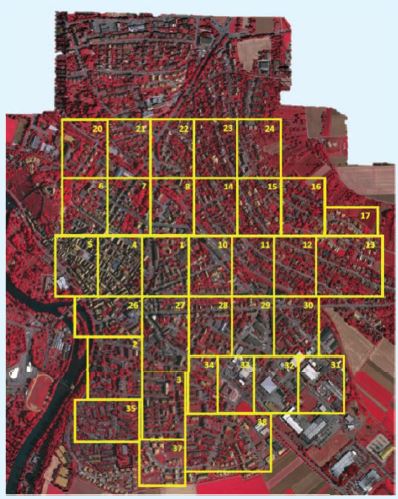}
  \caption{Illustration of 33 patches of ISPRS Vaihingen dataset (Source: \url{https://www.isprs.org/education/benchmarks/UrbanSemLab/2d-sem-label-vaihingen.aspx})} 
\label{fig:Vaihingen}
\end{figure}

{\it 2) ISPRS Vaihingen}\footnote{\url{https://www.isprs.org/education/benchmarks/UrbanSemLab/2d-sem-label-vaihingen.aspx}} is the earliest semantic segmentation dataset for identifying land cover classes in aerial images \cite{ISPRSVainighen} as shown in Fig. \ref{fig:evolution} (Tab.~\ref{tab:database}-\#220). This dataset contains a subset of images acquired by aerial cameras during the test of digital aerial cameras carried out by the German Association of Photogrammetry and Remote Sensing (DGPF) \cite{Cramer2010}. The dataset was prepared as part of a 2D semantic labeling contest organized by the International Society for Photogrammetry and Remote Sensing (ISPRS). The dataset contains images acquired over Vaihingen in Germany. In total, orthorectified images of varying sizes and a digital surface model (DSM) are provided for 33 patches covering the city of Vaihingen (Fig. \ref{fig:Vaihingen}). The ground sampling distance for the images and DSM is 9 cm. The three channels of the orthorectified images contain infrared, red, and green bands as acquired by the camera. The images are manually labeled for six common land cover classes: impervious surfaces, buildings, low vegetation, tree, car, and clutter/background.   

\subsection{Scene Classification}

\begin{figure*}
    \centering
    \includegraphics[width=\textwidth]{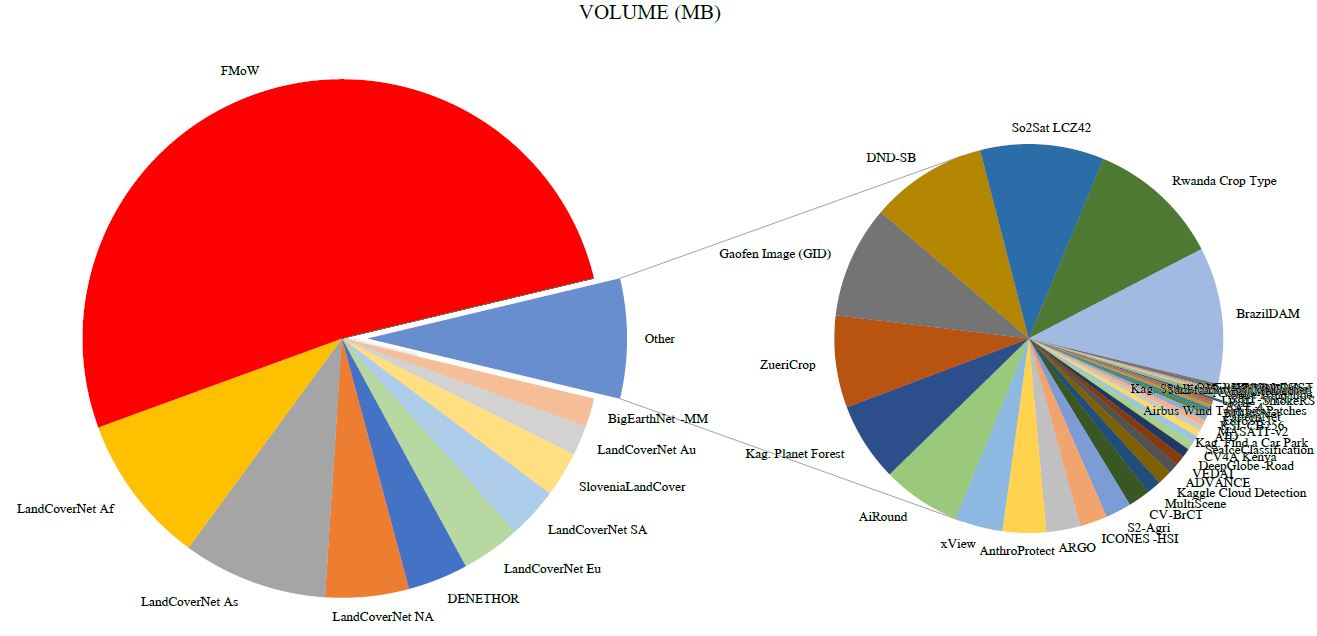}
    \caption{Relative volume distribution among datasets addressing scene classification.}
    \label{fig:data_class}
\end{figure*}

Remote sensing scene classification is closely related to semantic segmentation, and has been used in various application domains such as urban planning \cite{wang2019urban}, environment monitoring \cite{wen2021change}, LULC  estimation \cite{karpatne2016monitoring}, etc. The main difference is that instead of pixel-wise classification and resolution-preserving maps as output, in scene classification only one or more global labels are predicted for a given input image, aiming at a more comprehensive, generalized and context-aware understanding of the underlying scene. Similar to image classification, which has been the driving force behind the early days of deep learning in computer vision, research on remote sensing scene classification has led to the creation of more diverse and large-scale high-resolution image datasets. Fig.~\ref{fig:data_class} shows the relative volume distribution of remote sensing scene classification datasets. This section covers EuroSAT as one of the earliest and fMoW as one of the largest datasets (representing roughly 50\% of the available data in this task category), as well as BigEarthNet-MM that introduces an additional interesting aspect by providing multiple labels for each image.

\begin{figure}[ht]
  \centering
  \includegraphics[width=0.9\linewidth]{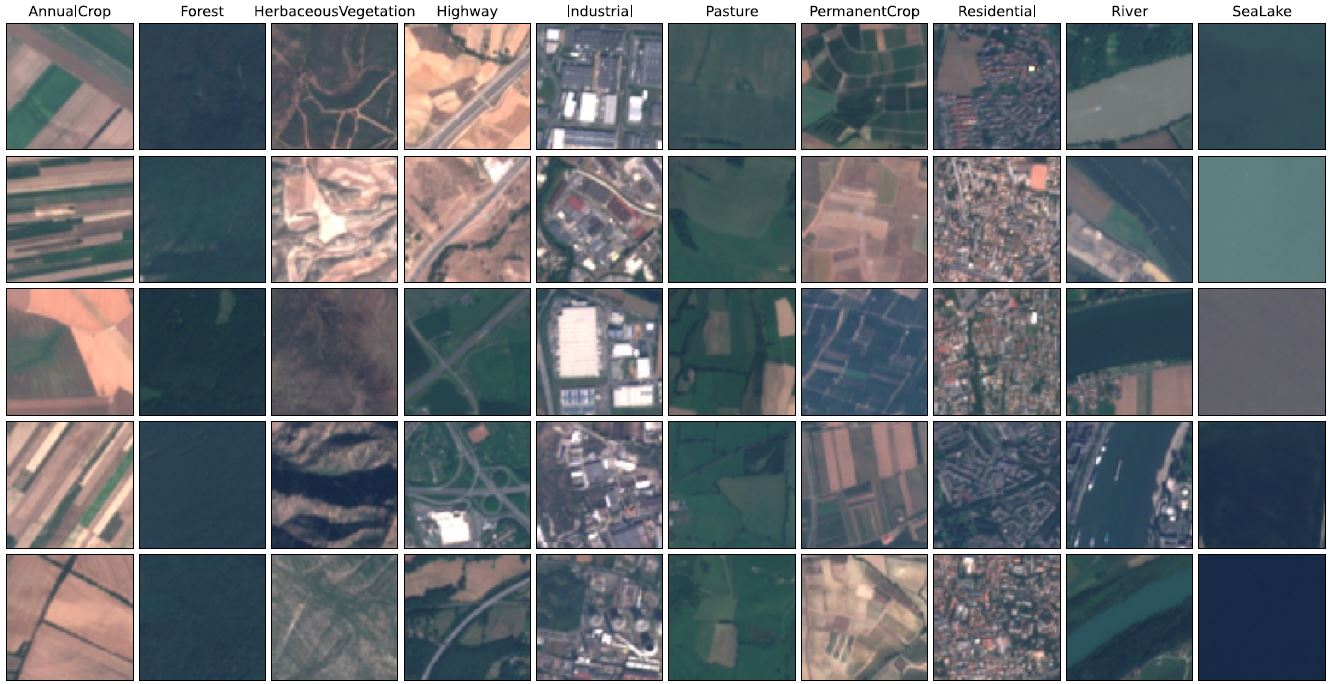}
\caption{Sample image patches of all ten classes covered in the EuroSAT dataset.}
\label{fig:EuroSat}
\end{figure}

{\it 1) EuroSAT}\footnote{\url{https://github.com/phelber/EuroSAT}} is one of the earliest large-scale datasets tailored for training deep neural networks for LULC classification tasks as Fig. \ref{fig:evolution} shows (Tab.~\ref{tab:database}-\#52). The dataset includes ten land cover classes (Fig. \ref{fig:EuroSat}), each containing 2,000-3,000 images, for 27,000 annotated and geo-referenced images with 13 different spectral bands of  $64 \times 64$ pixels. It contains multispectral images with a single label acquired from Sentinel-2 satellite images of cities in 34 European countries \cite{helber2019eurosat,helber2018introducing}. This dataset has been widely used for classification tasks. However, it may be used in a variety of real-world Earth observation applications, such as detecting changes in land use and land cover, as well as improving geographical maps as Sentinel-2 images are freely available \cite{helber2019eurosat,yamashkin2020improving,broni2021searching}. 

\begin{figure}[ht]
  \centering
  \includegraphics[width=0.9\linewidth]{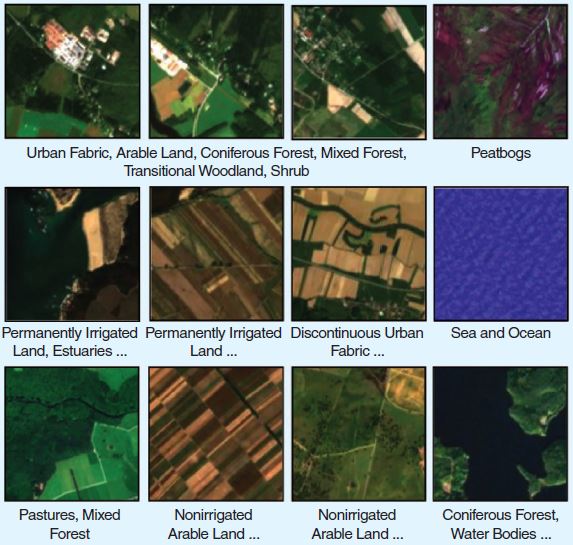}
  \caption{Sample image patches of several classes in the BigEarthNet-MM dataset with multiple labels being assigned to each image.}  
\label{fig:bigaerthnet}
\end{figure}

{\it 2) BigEarthNet-MM}\footnote{\url{https://bigearth.eu/}} (Tab.~\ref{tab:database}-\#40) is a benchmark archive that introduced an alternative nomenclature for images as compared to the traditional CORINE Land Cover (CLC) map for \textit{multi-label} image classification and image retrieval tasks. The CLC Level-3 nomenclature is arranged into 19 classes for semantically coherent and homogenous land cover classes. This archive is created using Sentinel-1 SAR and Sentinel-2 multispectral satellite images acquired between June 2017 and May 2018 over ten countries of Europe \cite{sumbul2021bigearthnet}. The first version of the BigEarthNet only included Sentinel-2 image patches. It has been augmented with Sentinel-1 image patches to form a multi-modal benchmark archive (also called BigEarthNet-MM). The archive comprises 590,326 image patches with 12 spectral and two polarimetric bands. As shown in Fig. \ref{fig:bigaerthnet}, each image patch is annotated with several land-cover classes (multi-labels). One of the key features of the BigEarthNet dataset is its large number of classes and images, which makes it suitable for training deep learning models. However, because it is a multi-label dataset, the complexity of the problem is significantly increased compared to single-label datasets \cite{chaudhuri2021interband}.
 
\begin{figure}[ht]
  \centering
  \includegraphics[width=0.9\linewidth]{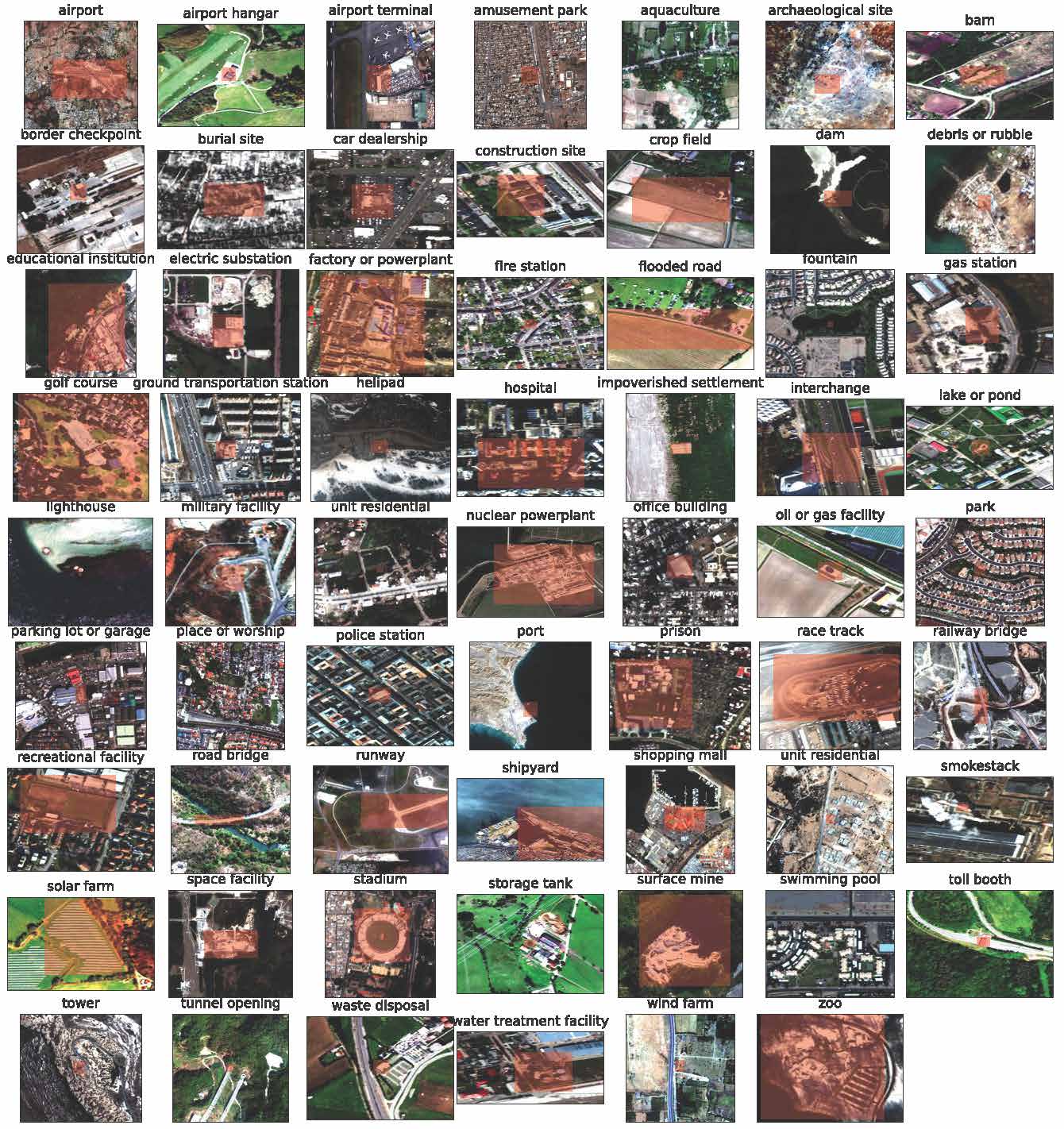}
  \caption{Sample image patches for several classes of the fMoW dataset.} 
\label{fig:fmowDataset}
\end{figure}

{\it 3) Functional Map of the World (fMoW)}\footnote{\url{https://github.com/fMoW/dataset}} (Tab.~\ref{tab:database}-\#76) is the largest dataset (Fig. \ref{fig:evolution}) for remote sensing scene classification in terms of the number of images \cite{christie2018functional}. The fMoW dataset is composed of approximately one million satellite images, along with metadata and multiple temporal views, as well as subsets of the dataset that are classified into 62 classes, which are used for training, evaluation, and testing. Each image in the dataset comes with one or multiple bounding boxes indicating the regions of interest. Some of these regions may be present in multiple images acquired at different times, adding a temporal dimension to the dataset. The fMoW dataset is available to the public in two image formats: the fMoW-Full and the fMoW-RGB. The fMoW-Full is in TIFF format and contains 4-band and 8-band multispectral imagery with a high spatial resolution resulting in 3.5TB of data, while the fMoW-RGB has a much smaller size of 200GB and includes all multispectral data converted to RGB in JPEG format. Examples of the classes in the dataset include flooded roads, military facilities, airstrips, oil and gas facilities, surface mines, tunnel openings, shipyards, ponds, and towers (see Fig. \ref{fig:fmowDataset} for some examples). The fMoW dataset has a number of important characteristics, such as global diversity, multiple images per scene captured at different times, multispectral imagery, and metadata linked to each image. 

\subsection{Object Detection}

\begin{figure*}
    \centering
    \includegraphics[width=\textwidth]{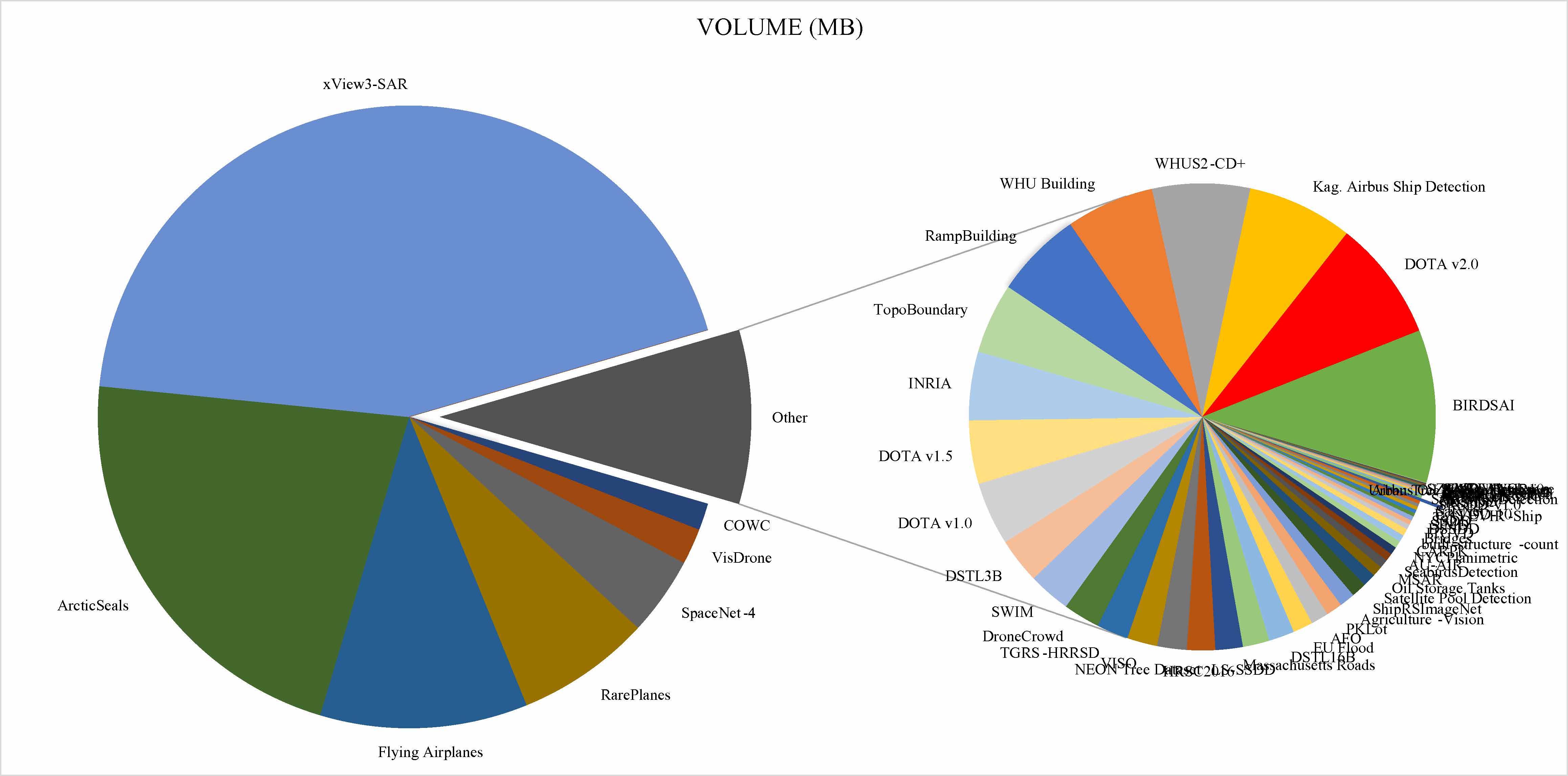}
    \caption{Relative volume distribution among datasets addressing object detection.}
    \label{fig:data_od}
\end{figure*}

The aim of object detection is to locate and identify the presence of one or more objects within an image, including objects with clear boundaries, such as vehicles, ships, and buildings, as well as those with more complex or irregular boundaries, for example LULC parcels \cite{Cheng2016ObjectDetection}. As seen in Fig.~\ref{fig:distribution}, object detection is one of the most widely studied research tasks.
Fig.~\ref{fig:data_od} shows the relative volume of the corresponding datasets. The xView3-SAR is by far the largest one. 
On the other hand, the family of DOTA datasets has a pioneering role in this field, placing them among the most popular datasets for object recognition tasks from remote sensing images. 

\begin{figure}[ht]
  \centering
  \includegraphics[width=0.9\linewidth]{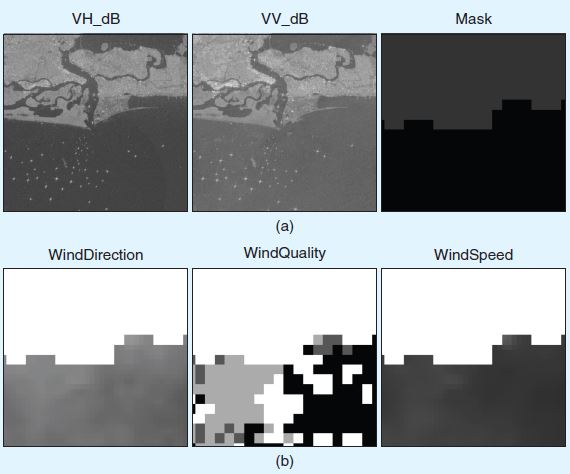}
  \caption{An example image stack of dual-polarimetric SAR images, water mask, and several wind properties from the xView3-SAR  dataset.} 
\label{fig:xView3-SAR}
\end{figure}

1) The xView3-SAR\footnote{\url{https://iuu.xview.us/}} Ship Detection dataset is the largest labeled dataset (Tab.~\ref{tab:database}-\#191), as Fig. \ref{fig:evolution} shows, for training machine learning models to detect and classify ships in SAR images. The dataset includes nearly 1,000 analysis-ready SAR images from the Sentinel-1 mission, which are annotated using a combination of automated and manual analysis (see Fig. \ref{fig:xView3-SAR} for some image samples). The images are accompanied by co-located bathymetry and wind state rasters, and the dataset is intended to support the development and evaluation of machine learning approaches for detecting ''dark vessels'' not visible in conventional monitoring systems.

\begin{figure}[ht]
  \centering
  \includegraphics[width=0.9\linewidth]{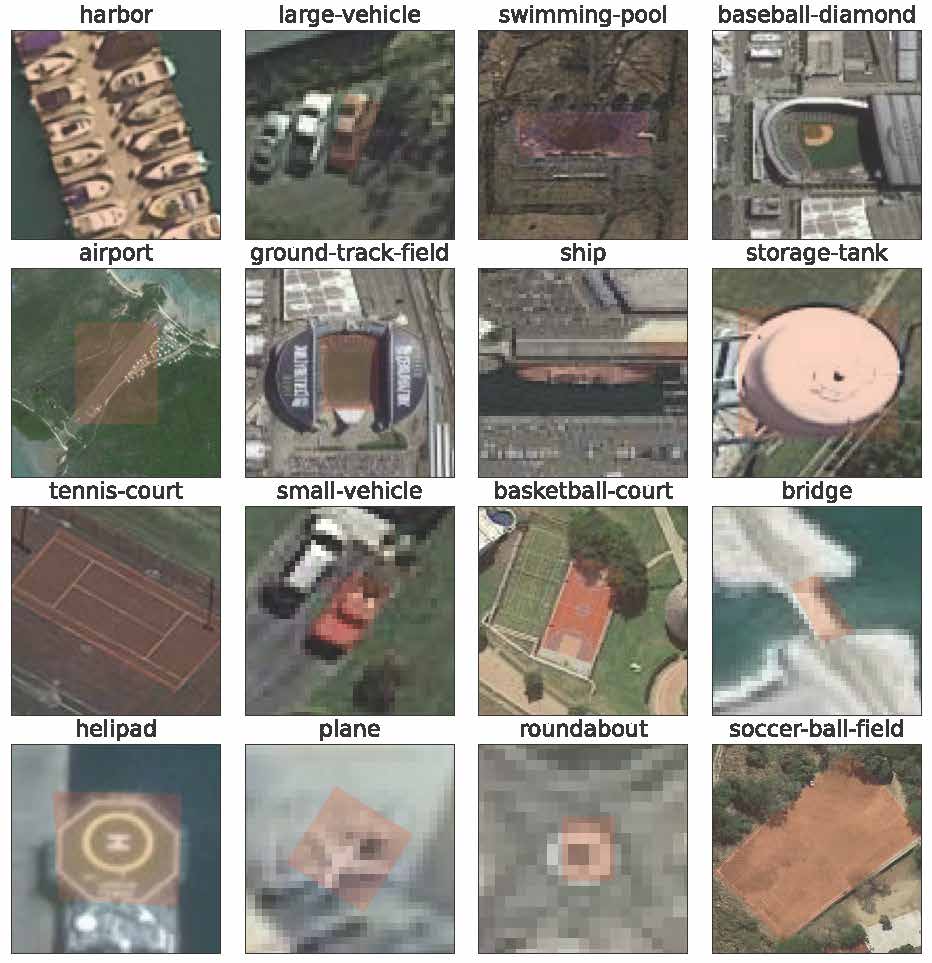}
  \caption{Several example image samples of the DOTA dataset.} 
\label{fig:DOTA_sample}
\end{figure}

2) {\it DOTA}\footnote{\url{https://captain-whu.github.io/DOTA/}} is one of the most popular and largest object detection datasets (Tab.~\ref{tab:database}-\#109/114) in terms of labeled object instances. It includes 2,806 images acquired from Google Earth and the China Center for Resources Satellite Data and Application \cite{Xia_2018_CVPR,Ding_2019_CVPR,9560031}. The DOTA dataset is available in three different versions: DOTA-v1.0, DOTA-v1.5, and DOTA-v2.0. The image size in the initial version ranges from $800 \times  800$ pixels to $4000 \times 4000$ pixels, with 188,282 object instances with various scales and angular orientations and a total of 15 object categories. DOTA-v1.5 adds various small objects (less than 10 pixels) and a new container crane category with 402,089 instances, whereas DOTA-v2.0 adds two categories, airport and helipad, with 11,268 images and 1,793,658 instances, respectively. Some image samples of the DOTA dataset are presented in Fig.~\ref{fig:DOTA_sample}.

\subsection{Change Detection}

\begin{figure*}
    \centering
    \includegraphics[width=\textwidth]{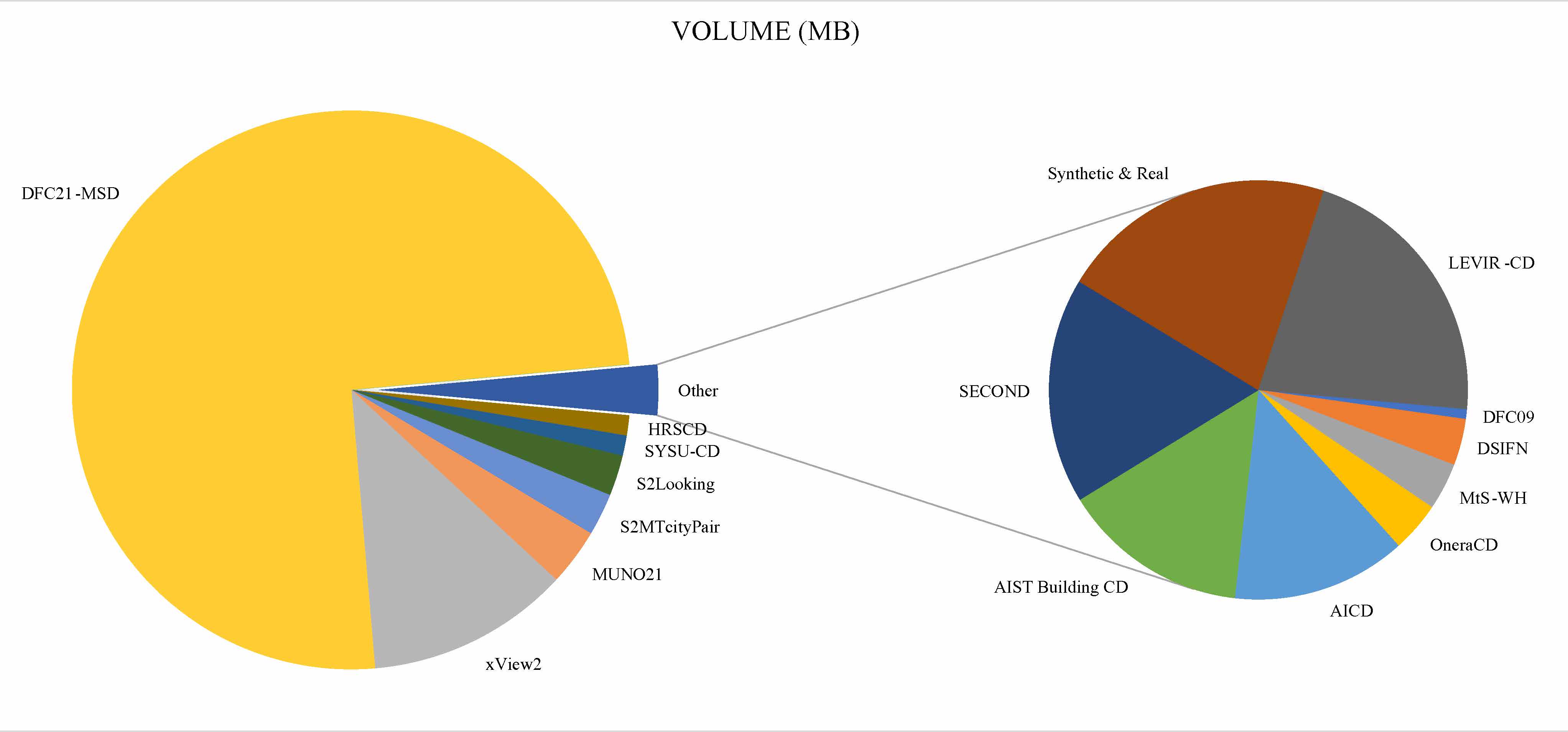}
    \caption{Relative volume distribution among datasets addressing change detection.}
    \label{fig:data_cd}
\end{figure*}

Change detection (CD) in remote sensing aims to identify temporal changes by analyzing multi-temporal satellite images of the same location. CD is a popular task in Earth observation, as it fosters monitoring environmental changes through artificial or natural phenomena. Fig.~\ref{fig:distribution} shows that the number of dedicated change detection datasets is small compared to other applications. 
Fig.~\ref{fig:data_cd} shows that the available data is dominated by the DFC20 dataset (Track MSD) which focuses on semantic change detection, followed by xView2 tackling building damage assessment in the context of natural disasters. 
We chose {\it LEVIR-CD} as a recent dataset example and the {\it Onera Satellite Change Detection Dataset} as one of the first large-scale datasets.

\begin{figure}[ht]
  \centering
  \includegraphics[width=0.9\linewidth]{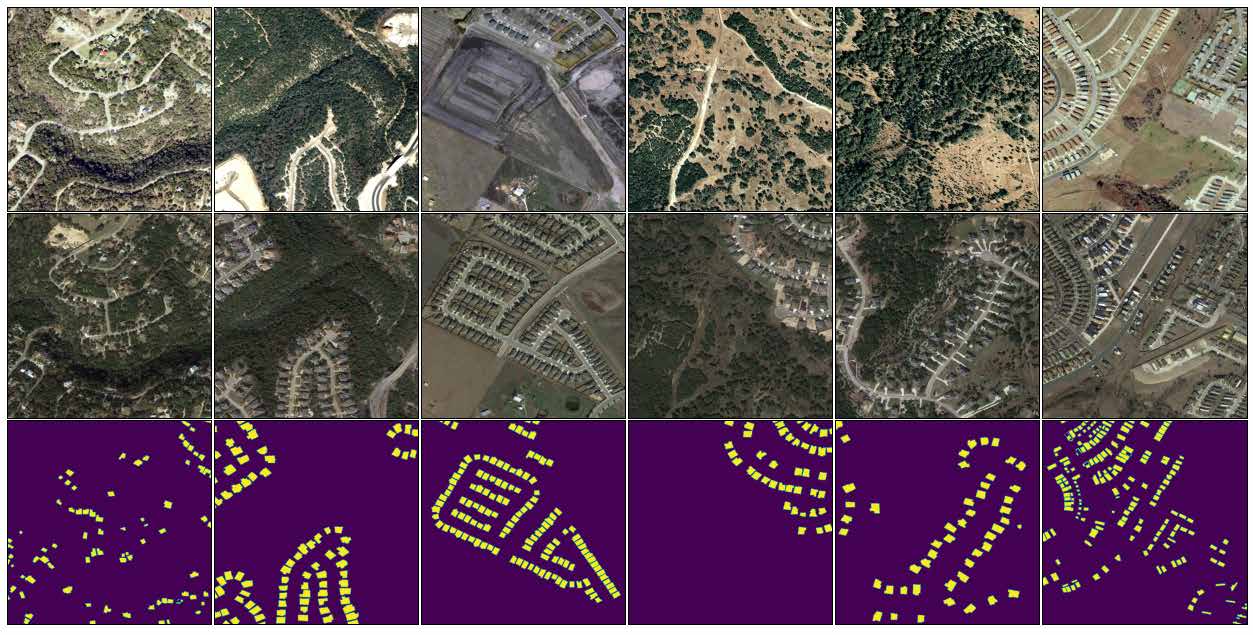}
  \caption{Examples of annotated samples in the LEVIR-CD dataset.} 
\label{fig:levircd}
\end{figure}

{\it 1) {LEVIR-CD}}\footnote{\url{https://justchenhao.github.io/LEVIR/}}  is one of the most recent and largest change detection datasets as seen in Fig. \ref{fig:evolution} (Tab.~\ref{tab:database}-\#15). It is mainly developed for the evaluation of deep learning algorithms on building-related changes, including building growth (the transition from soil/grass/hardened ground, building under construction to new build-up regions) and building decline \cite{Chen2020}. The dataset comprises 637 optical image patch pairings extracted from Google Earth (GE) with a resolution of $1024 \times 1024$ pixels, acquired over a time span of 5 to 14 years. LEVIR-CD covers various structures, including villa residences, tall apartments, small garages, and large warehouses, from 20 distinct regions in multiple Texas cities. The fully annotated LEVIR-CD dataset comprises 31,333 distinct change-building instances, some of which are illustrated in Fig. \ref{fig:levircd} generated from the bi-temporal images by remote sensing image interpretation specialists.

\begin{figure}[ht]
  \centering
  \includegraphics[width=0.9\linewidth]{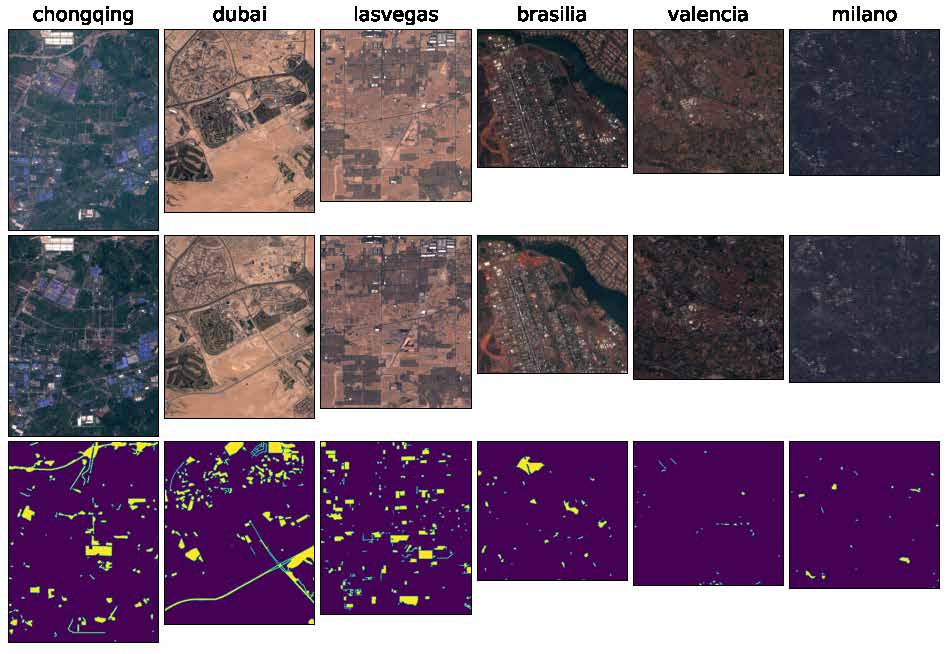}
  \caption{Examples of annotated samples in the Onera Satellite Change Detection dataset.} 
\label{fig:onera}
\end{figure}

{\it 2) Onera Satellite Change Detection}\footnote{\url{https://rcdaudt.github.io/oscd/}} is one of  the first, larger change detection datasets, as Fig \ref{fig:evolution} shows (Tab.~\ref{tab:database}-\#11), containing multispectral image pairs from Sentinel-2. This dataset includes 24 pairs of multispectral images acquired from Sentinel-2 satellites between 2015 and 2018, focusing on urban changes such as new buildings or roads \cite{daudt2018urban}. The locations are selected from around the world, including Brazil, the United States, Europe, the Middle East, and Asia. The reference data for pixel-level change is provided for all 14 training and 10 test image pairs. As illustrated in Fig.~\ref{fig:onera}, the annotated changes are primarily associated with urban changes, such as new buildings or roads.

\subsection{Super Resolution/Pansharpening}

Pansharpening is one of the oldest data fusion approaches in remote sensing and aims to increase the spatial resolution of a multispectral image by combining it with a panchromatic image. Due to the resolution difference between panchromatic and multispectral sensors, pansharpening is an exciting topic in remote sensing as it can provide a means to obtain higher-resolution data without better sensor technology. We select the {\it Proba-V}, {\it PAirMax}, and {\it WorldStrat} datasets as examples to showcase the peculiarities of datasets designed for that particular application.

\begin{figure}[ht]
  \centering
  \includegraphics[width=0.9\linewidth]{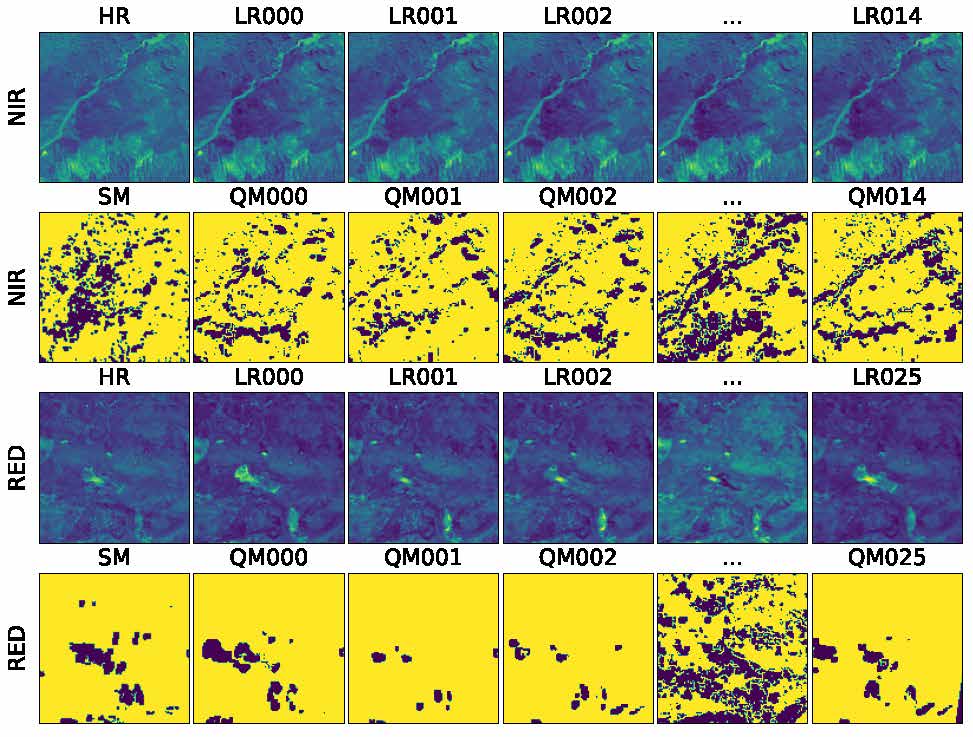}
\caption{Sample images of Proba-V dataset. Each sample consists of one high resolution (HR) and several low resolution (LR) images, each with a quality map (QM) showing which pixels are concealed (e.g. through clouds, etc.).}
\label{fig:Proba-V}
\end{figure}

{\it 1) {Proba-V}}\footnote{\url{https://kelvins.esa.int/proba-v-super-resolution/data/}} is the earliest dataset available for super-resolution  \cite{ProbaV2019} as Fig. \ref{fig:evolution} shows (Tab.~\ref{tab:database}-\#318). This dataset contains radiometrically and geometrically corrected Top-Of-Atmosphere reflectances in Plate Carre projection from the PROBA-V mission of the European Space Agency (Fig. \ref{fig:Proba-V}). The RED and NIR spectral band data at 300~m and 100~m resolution are collected for 74 selected regions across the globe. Super-resolution might be affected by the presence of pixels with clouds, cloud shadows, and ice/snow cover. Therefore, this dataset contains a quality map indicating pixels affected by clouds that should not be considered for super-resolution. The dataset contains one 100~m resolution image and several 300~m resolution images of the same scene. 

\begin{figure}[ht]
  \centering
  \includegraphics[width=0.9\linewidth]{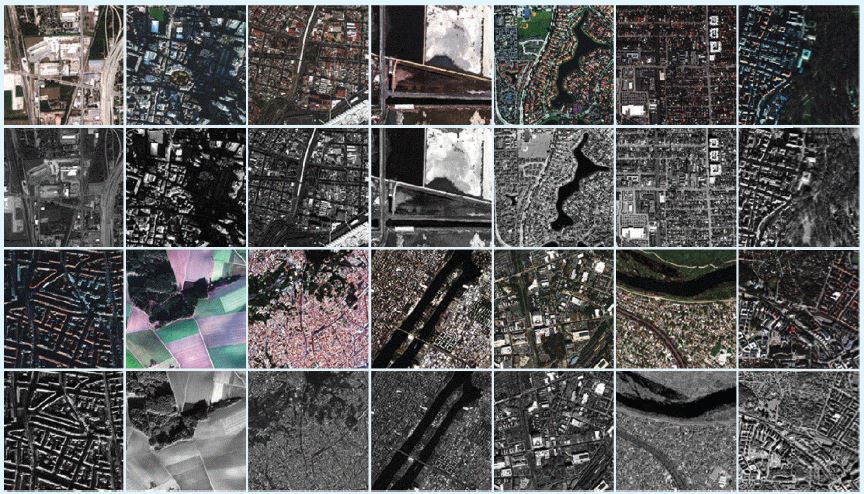}
  \caption{Sample images of PAirMax dataset (Source: \url{https://perscido.univ-grenoble-alpes.fr/datasets/DS353})} 
\label{fig:pairmax}
\end{figure}

{\it 2) {PAirMax}}\footnote{\url{https://perscido.univ-grenoble-alpes.fr/datasets/DS353}}  is a recently published (Tab.~\ref{tab:database}-\#319/324) yet popular dataset for pansharpening. It contains 14 multispectral and panchromatic image pairs from six sensors onboard satellite constellations of Maxar Technologies and Airbus \cite{Vivone2021}. As seen in Fig. \ref{fig:pairmax}, most images are acquired over urban areas, highlighting several challenges for pan-sharpening (such as high contrast features and adjacent regions with different spectral features). The work \cite{Vivone2021} also discusses the best practices to be followed in preparing high-quality full-resolution and reduced-resolution images for pan-sharpening applications. In this dataset, the panchromatic band has $4 \times 4$ times more pixels than the multispectral bands. The multispectral bands are near the visible infra-red region. The dataset also contains nine reduced-resolution test cases per Wald’s protocol. 

\begin{figure}[ht]
  \centering
  \includegraphics[width=0.9\linewidth]{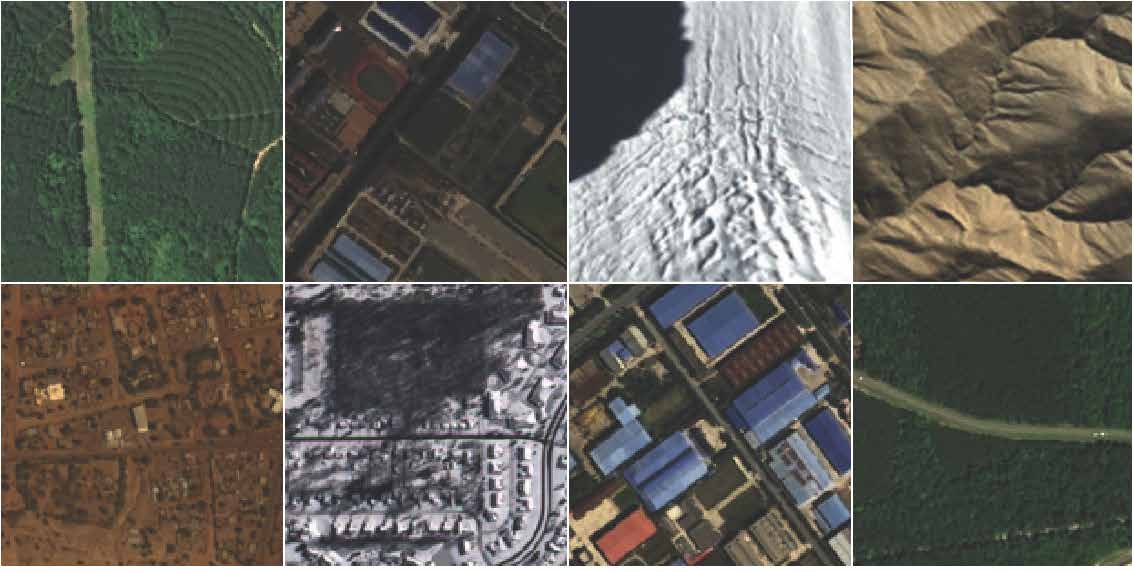}
  \caption{Sample images of WorldStrat dataset (Source:  \cite{cornebise2022open}) }
\label{fig:WorldStrat}
\end{figure}

{\it 3) {WorldStrat}}\footnote{\url{https://worldstrat.github.io/}}   is a recently introduced dataset for super-resolution \cite{cornebise2022open} and the largest in terms of volume (107 GB) as Fig.\ref{fig:evolution} shows (Tab.~\ref{tab:database}-\#325). This dataset contains high-resolution images from Airbus 6/7 along with temporally matched 16 low-resolution images from Sentinel-2 satellites. The high resolution images are over five spectral bands: Panchromatic band at 1.5 m/pixel resolution and Red, Green, Blue, and Near Infrared bands at 6 m/pixel. The low resolution ranges from 10 m/pixel to 60 m/pixel (Figure~\ref{fig:WorldStrat}). In total, the dataset covers an area of around 10,000 km$^2$ and attempts to represent all types of land-use across the world.  Notably, the dataset contains non-settlement and under-represented locations such as illegal mining sites, settlements of persons at risk, etc.

\section{Working with Remote Sensing Datasets}
\label{sec:practical}

This section provides guidance on how to leverage available datasets to their full potential. Two of the main difficulties caused by information asymmetry (i.e. the information imbalance between the data providers and the data users) \cite{Schmitt2021} are finding suitable datasets and easy prototyping of machine learning approaches using such datasets. Here we discuss resources to gain an overview of existing datasets and download actual data, but also provide examples of Earth observation-oriented machine learning programming libraries.

\subsection{Data Availability}\label{sec:data}

Data availability covers two distinct aspects: On the one hand, access to the curated benchmark datasets, i.e. how such datasets are made available to the public. This section provides several examples of the most common data sources below. 
On the other hand, the actual non-curated measurements as acquired by the different sensors such as satellites, planes, and UAVs are often available, too. Many data providers offer either their complete database for public access (e.g. the European Copernicus Open Access Hub\footnote{\url{https://scihub.copernicus.eu/}}) or at least parts of their image acquisitions (e.g. via Open Data Programs such as those from Maxar\footnote{\url{https://www.maxar.com/open-data}} and Capella Space\footnote{\url{https://registry.opendata.aws/capella_opendata/}} or through scientific proposals as possible for TerraSAR-X/TanDEM-X data\footnote{\url{https://sss.terrasar-x.dlr.de/}}). 
In principle, these data sources are highly valuable as they offer free access to petabytes of EO data that can either be used to compile benchmark datasets by augmenting it with reference data for a specific task or being leveraged for modern training paradigms such as self-supervised learning. 
An important aspect that is unfortunately sometimes ignored is the licensing of the corresponding data products: While direct usage (at least for scientific purposes) is usually free, redistribution of this data is often prohibited. Nevertheless, such image data finds its way into public benchmark datasets essentially causing copyright infringements. 
Additionally to being in direct conflict with scientific integrity, such behavior is likely to be less tolerated in the future given the rising commercial value of EO imagery. 
Creators of future benchmark datasets should be fully aware of the license under which the leveraged EO data is accessible and how it limits its use in public datasets. 
In parallel, data providers should consider a licensing scheme that allows non-commercial redistribution for scientific / educational purposes to further facilitate the development and evaluation of modern machine-learning techniques.

\begin{figure}[ht]
  \centering
  \includegraphics[width=0.9\linewidth]{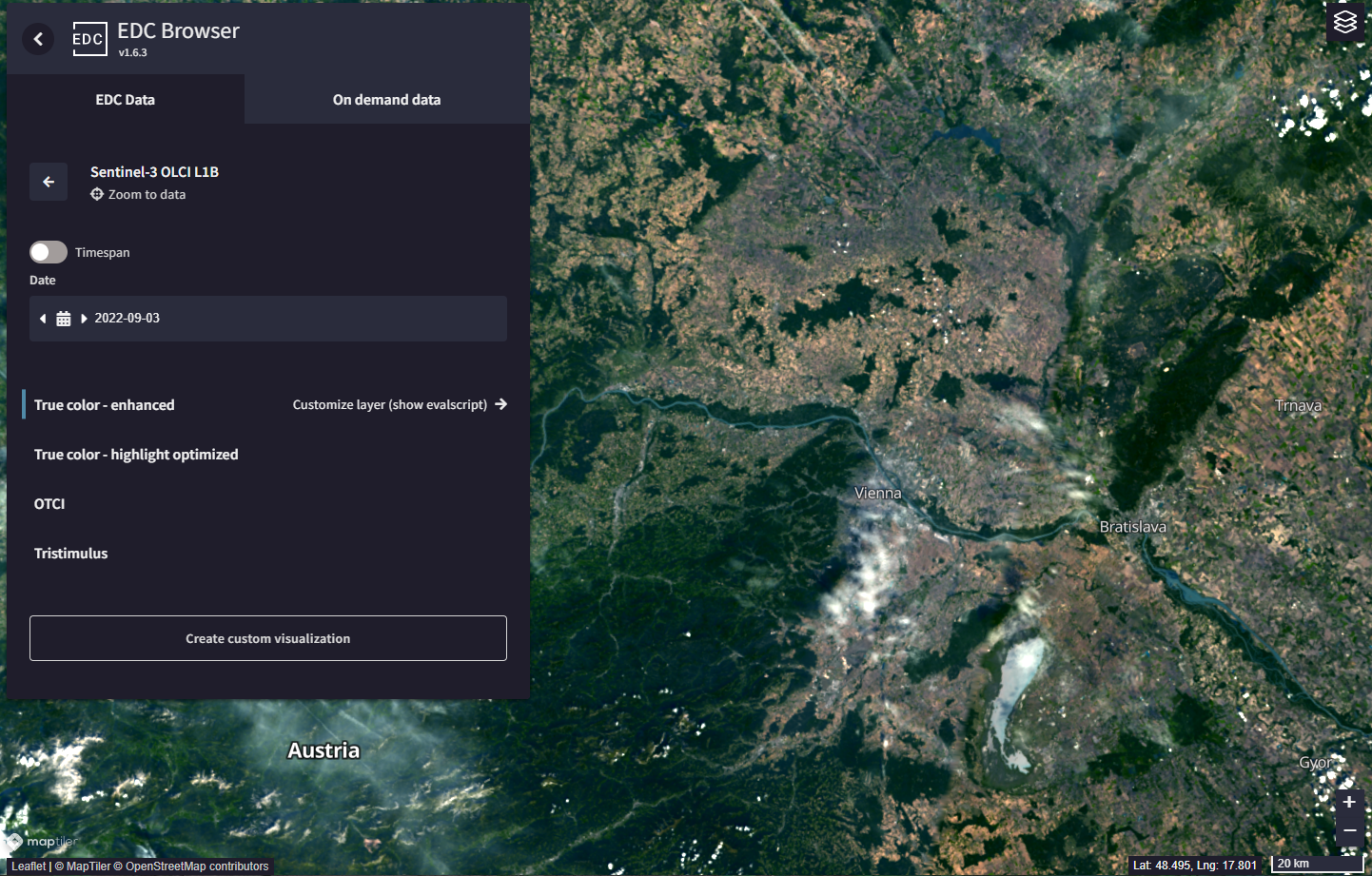}
  \caption{Illustration of the EDC Browser, where users can easily search and order satellite data (Source: \url{https://browser.eurodatacube.com/}).}
\label{fig:EDC_demo}
\end{figure}

1) \textit{IEEE DataPort}\footnote{\url{https://ieee-dataport.org}} is a valuable and easily accessible data platform that enables users to store, search, access, and manage data. The platform is designed to accept all formats and sizes of datasets (up to 2TB), and it provides both downloading capabilities and access to datasets in the Cloud. Both individuals and institutions can indefinitely store and make datasets easily accessible to a broad set of researchers, engineers, and industry. In particular, most of the datasets used for the past IEEE GRSS Data Fusion Contests have been published on this platform. However, unless being a dataset associated with a competition or being submitted as open access, an IEEE account is required to access and download.

2) \textit{Radiant MLHub}\footnote{\url{https://mlhub.earth}} enables users to access, store, register, and share high-quality open datasets for training ML models in EO. It’s designed to encourage widespread collaboration and the development of trustworthy applications. The available datasets in this platform cover research topics like crop type classification, flood detection, tropical storm estimation, etc.

3) \textit{Euro Data Cube (EDC)}\footnote{\url{https://eurodatacube.com}} provides a global archive of analysis-ready data (Sentinel, Landsat, MODIS, etc.), where users can search and order data using the graphical interface EDC Browser (see Fig.~\ref{fig:EDC_demo}). It also enables users to store, analyze, and distribute user-contributed data content with simple APIs. 

\begin{figure*}
  \centering
  \includegraphics[width=0.9\linewidth]{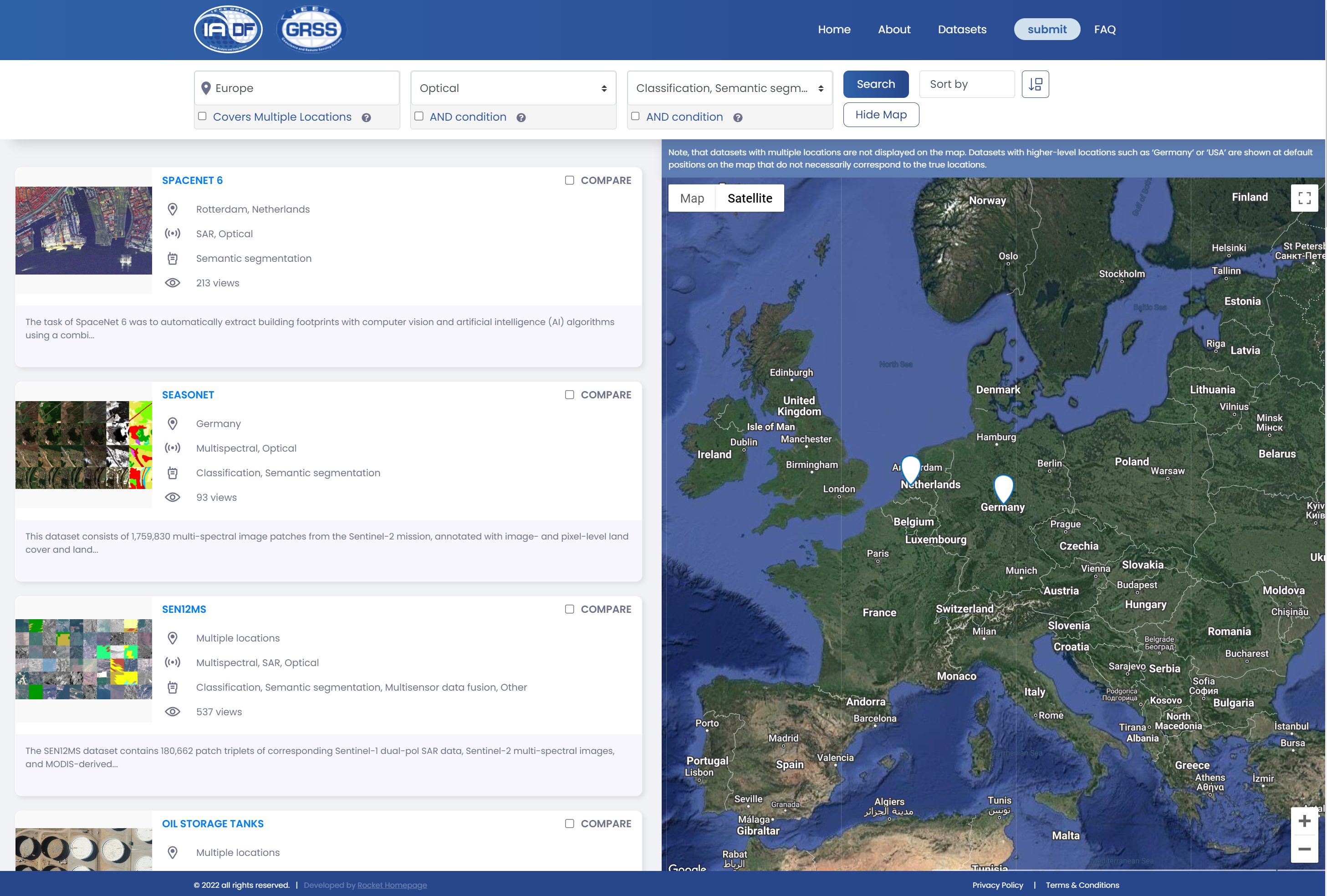}
  \caption{Illustration of the map view of the EOD data catalog. }
\label{fig:EOD_demo}
\end{figure*}

4) \textit{NASA Earthdata}\footnote{\url{https://www.earthdata.nasa.gov}} provides access to a wide range of remote sensing datasets, including satellite imagery, atmospheric data, and land cover data, and offers a number of programming APIs. For example, the Application for Extracting and Exploring Analysis Ready Samples (AppEEARS) API enables users to perform data access and transformation processes easily and efficiently.

5) \textit{Maxar Open Data Program}\footnote{\url{https://www.maxar.com/open-data}} provides access to high-resolution remote sensing imagery collected since 2017, amounting to a total area of 1,834,152 square kilometers. It also features several programming APIs, such as the Maxar Catalog API and the Analysis-Ready Data API. In addition, this program seeks to assist the humanitarian community by furnishing essential and useful information to support response efforts when crises occur.

6) \textit{OpenAerialMap}\footnote{\url{https://openaerialmap.org}} is a community-driven platform that provides access to openly licensed satellite and unmanned aerial vehicle imagery from across the globe, accompanied by programming APIs for data access and processing. The platform currently hosts 12,614 images captured by 721 sensors, and all images can be traced in OpenStreetMap\footnote{\url{https://openstreetmap.org}}.

7) \textit{EarthNets}\footnote{\url{https://earthnets.nicepage.io}} is an interactive data-sharing platform with more than 400 publicly available datasets in the geoscience and remote sensing field, which covers essential EO tasks like land use/cover classification, change/disaster monitoring, scene understanding, climate change, and weather forecasting \cite{xiong2022earthnets}. Each benchmark dataset provides detailed meta information like spatial resolution and volume. Besides, it also supports standard data loaders and codebases for different remote sensing tasks, which enables users to conduct a fair and consistent evaluation of deep learning methods on the benchmark datasets.

8) \textit{Earth Observation Database (EOD)}\footnote{\url{https://eod-grss-ieee.com}} is an interactive online platform for cataloging different types of datasets leveraging remote sensing imagery, which is developed by IEEE GRSS Image Analysis and Data Fusion (IADF) Technical Committee \cite{schmitt2022eod}. The key feature of EOD is to build a central catalog that provides an exhaustive list of available datasets with their basic information, which can be accessed and extended by the community and queried in a structured and interactive manner (see Fig.~\ref{fig:EOD_demo} for an example of the user interface). For more information, please refer to the corresponding box.

\noindent\fbox{%
    \parbox{\linewidth}{%
    \textbf{The IEEE-GRSS Earth Observation Database}\\
       The core of the EOD is a database of user-submitted datasets. It can be searched by a combination of various queries, including location, as well as different combinations of sensors and tasks. Search results (as well as the whole database) can be viewed either in an illustrated list view or in an interactive map view indicating those datasets that cover only a specific location with a marker (see Fig.~\ref{fig:EOD_demo}). Clicking on one of the markers in the map view will limit the list to datasets at this specific location. Clicking on a dataset in the list will open a new window displaying detailed information about this dataset which includes
    
    \begin{itemize}
        \item Geographic location;
        \item Sensor modality;
        \item Task/application;
        \item Data size;
        \item URL to access the data;
        \item Number of views;
        \item A graphical representation;
        \item A brief description of the dataset.
    \end{itemize}
    A helpful function is the \textit{compare} option, which allows a direct side-to-side comparison of this information from two or more datasets in a newly opened window.
    EOD is meant as a community-driven catalog maintained by GRSS IADF, i.e. adding datasets is neither limited to IADF members nor to the creators of the dataset but anybody with an interest that the visibility and accessibility of a certain dataset is increased can request to include it into EOD. The submission requests are reviewed by IADF for completeness and correctness before the dataset is added to the database and visible to the public.
}}%

\subsection{EO-oriented ML Libraries}

Most existing ML libraries are developed for classic computer vision tasks, where the input image is usually single-channel (grayscale) or with RGB bands. EO datasets are often of large volumes with highly diverse data types, different numbers of spectral bands, and spatial resolutions, as illustrated in Fig.~\ref{fig:sizemeasure}. 
The code base for processing such data samples is often highly complex and difficult to maintain. One approach to increase readability, reusability, and maintainability is to modularize the code and encapsulating different tasks by decoupling processing the dataset (data loaders, visualization, pre-processing, etc.) and applying machine learning models (training, prediction, model selection, evaluation, etc.). A major challenge in training advanced ML models for EO tasks is the implementation of an easy-to-use, yet efficient data loader explicitly designed for geoscience data that loads and preprocesses a complex dataset and produces an iterable list of data samples in a customizable way. 

This section introduces several well-known packages designed explicitly for geoscience and remote sensing data. Since PyTorch \cite{paszke2019pytorch} and TensorFlow \cite{abadi2016tensorflow} (note that Keras is now a part of TensorFlow) are the most widely used deep learning frameworks, we mainly focus on the introduction for the existing libraries using these two deep learning frameworks as the backend.

\begin{figure}
  \centering
  \includegraphics[width=0.9\linewidth]{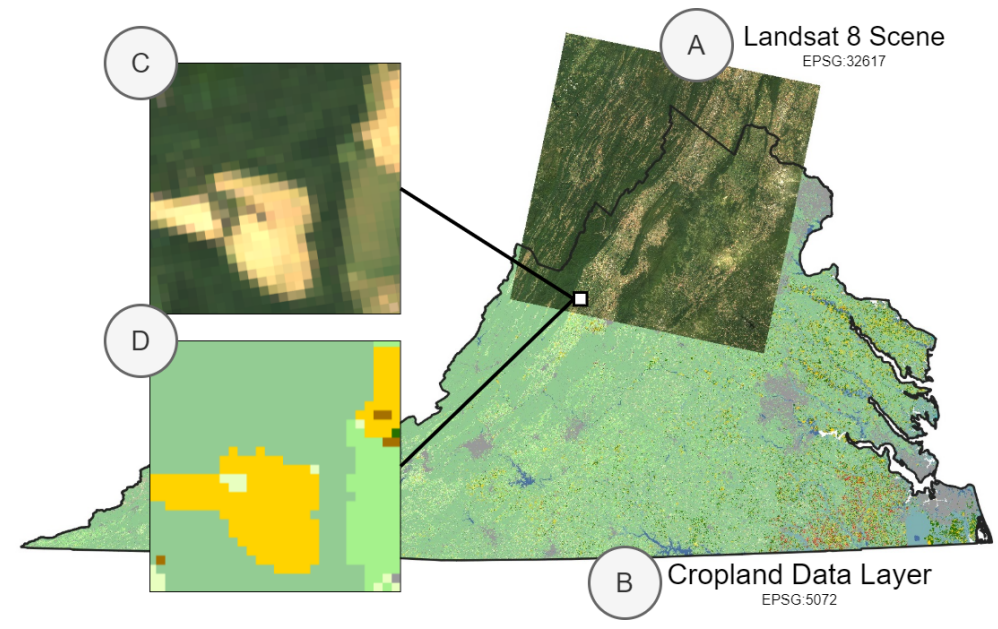}
  \caption{Illustration of sampling from heterogeneous geospatial data layers (Source: \cite{stewart2021torchgeo}). With the TorchGeo package, users can directly focus on training ML models without manually preprocessing the heterogeneous geospatial data, including aligning data layers by reprojection and resampling.}
\label{fig:torchgeo}
\end{figure}

1) \textit{TorchGeo}\footnote{\url{https://github.com/microsoft/torchgeo}} is an open-source PyTorch-based library, which provides datasets, samplers, transforms, and pre-trained models specific to geospatial data \cite{stewart2021torchgeo}. The main goal of this library is to simplify the process of interacting with complex geospatial data and make it easier for researchers to train ML models for EO tasks. Fig.~\ref{fig:torchgeo} provides an example of sampling pixel-aligned patch data from heterogeneous geospatial data layers using the TorchGeo package. Since different layers usually have different coordinate systems and spatial resolutions, patches sampled from these layers in the same area may not be pixel-aligned. Therefore, in a practical application scenario researchers need to conduct a series of preprocessing operations such as reprojecting and resampling of the geospatial data before training ML models, which is time-consuming and laborious. To address this challenge, TorchGeo provides data loaders tailored for geospatial data, which support transparently loading data from heterogeneous geospatial data layers with relatively simple code (an example is shown below).

\lstset{language=Python}
\begin{lstlisting}
from torch.utils.data import DataLoader
from torchgeo.datasets import CDL, Landsat8, stack_samples
from torchgeo.samplers import RandomGeoSampler

# Loading Landsat8 and CDL layers
landsat8 = Landsat8(root="...")
cdl = CDL(root="...", download=True, checksum=True)

# Take the intersection of Landsat8 and CDL
dataset = landsat8 & cdl

# Sample 10,000 256 x 256 image patches
sampler = RandomGeoSampler(dataset, size=256, length=10000)

# Build a normal PyTorch DataLoader with the sampler
dataloader = DataLoader(dataset, batch_size=128, sampler=sampler, collate_fn=stack_samples)

for batch in dataloader:
    image = batch["image"]
    mask = batch["mask"]

    # Train a model
    ...
\end{lstlisting}

A more detailed introduction about the supported geospatial datasets in TorchGeo can be found in \cite{stewart2021torchgeo}. 

\begin{figure}
  \centering
  \includegraphics[width=0.9\linewidth]{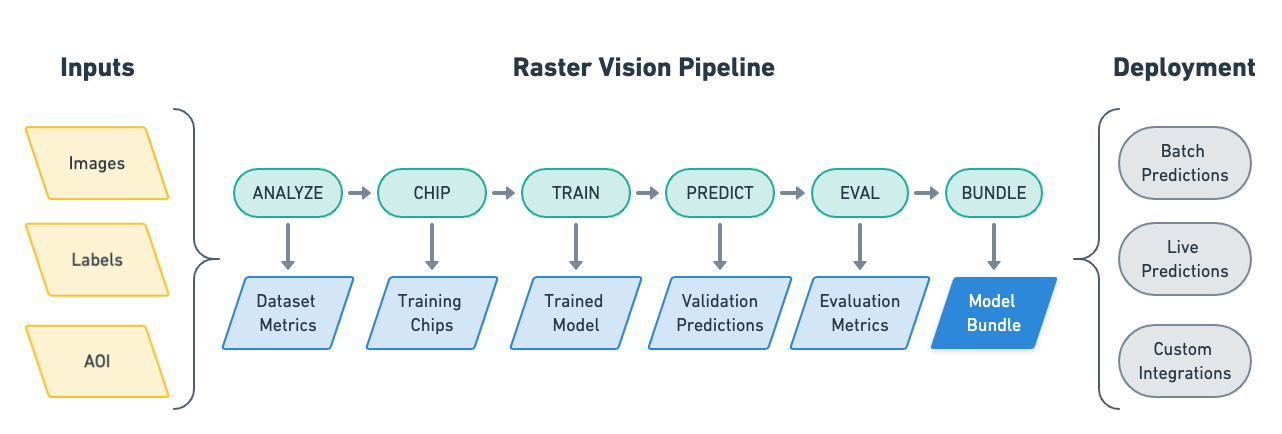}
  \caption{Illustration of the pipeline of the RasterVision package (Source: \url{https://github.com/azavea/raster-vision}).}
\label{fig:rastervision}
\end{figure}

2) \textit{RasterVision}\footnote{\url{https://github.com/azavea/raster-vision}} is an open-source Python framework that aims to simplify the procedure in building deep learning-based computer vision models on satellite, aerial, and other types of geospatial data (including oblique drone imagery). It enables users to efficiently construct a deep learning pipeline, including training data preparation, model training, model evaluation, and model deployment, without any expert knowledge of ML. Specifically, RasterVision supports chip classification, object detection, and semantic segmentation with the PyTorch backend on both CPUs and GPUs with built-in support for running in the cloud using AWS. The framework is extensible to new data sources, tasks (e.g., instance segmentation), backends (e.g., Detectron2), and cloud providers. Fig.~\ref{fig:rastervision} provides the pipeline of the RasterVision package. A more comprehensive tutorial for this package can be found at \url{https://docs.rastervision.io/}.

3) \textit{Keras Spatial}\footnote{\url{https://github.com/IllinoisStateGeologicalSurvey/keras-spatial}} provides data samplers and tools designed to simplify the preprocessing of geospatial data for deep learning applications with the Keras backend. It provides a data generator that reads samples directly from raster layers without creating small patch files before running the model. It supports loading raster data from local files or cloud services. Necessary preprocessing, like reprojecting and resampling, is also conducted automatically. Keras Spatial supports sample augmentation using a user-defined callback system to improve the flexibility of data management. A simple demo code using the \texttt{SpatialDataGenerator} class from Keras Spatial to prepare a training set for a deep learning model is given below:

\lstset{language=Python}
\begin{lstlisting}
from keras_spatial import SpatialDataGenerator

# Loading labels from a local file
labels = SpatialDataGenerator()
labels.source = '/path/to/labels.tif'
# Sample 128 x 128 patches
labels.width, labels.height = 128, 128
# Set a geodataframe with 200 x 200 
# in projection units of the original raster
df = labels.regular_grid(200,200)

# Loading images from a file on the cloud
samples = SpatialDataGenerator()
samples.source = 'https://server.com/files/data.tif'
samples.width, samples.height = labels.width, label.height

# The training set generator
train_gen = zip(labels.flow_from_dataframe(df), patches.flow_from_dataframe(df))

# Train a model
model(train_gen)
\end{lstlisting}

\subsection{Geospatial Computing Platforms}
In addition to ML libraries, public geospatial computing platforms, such as Google Earth Engine (GEE), for EO tasks in practical application scenarios offer a series of benefits including:
\begin{itemize}
    \item \textit{Access to large-scale datasets}: Geospatial computing platforms usually provide access to large and diverse geospatial datasets, such as satellite imagery, weather data, and terrain data. These datasets may be time-consuming and expensive to acquire on one's own but can be easily accessed through a geospatial computing platform using the cloud service. 
    \item \textit{Scalability}: Geospatial computing platforms are designed to address large-scale geospatial data processing and analysis. They usually provide cloud-based computing resources that can be easily scaled up or down to meet researchers' needs. This makes it easier to perform complex geospatial analysis that would be difficult to do on local machines considering the limited computing resources.
    \item \textit{Pre-built tools and APIs}: Geospatial computing platforms usually provide pre-built tools and programming APIs for image processing, geocoding, and data visualization, making it much easier for researchers to work with geospatial data.
\end{itemize}
Several representative geospatial computing platforms are listed below.

1) \textit{Google Earth Engine (GEE)}\footnote{\url{https://earthengine.google.com}} is a cloud-based platform designed for large-scale geospatial data analysis and processing. It provides a range of tools and APIs that allow users to analyze and visualize geospatial data, including raster and vector processing tools, machine learning algorithms, and geospatial modeling tools. In addition, GEE provides access to powerful computing resources, such as virtual machines and storage, to enable users to perform complex geospatial analyses. The GEE Data Catalog contains over forty years of historical imagery and scientific datasets for Earth science, which are updated and expanded daily. These datasets cover various topics such as climate, weather, surface temperature, terrain, and land cover. Notable datasets available on the GEE Data Catalog include Planet SkySat Public Ortho Imagery (collected for crisis response events) \cite{tamiminia2020google} and NAIP\footnote{\url{https://developers.google.com/earth-engine/datasets/catalog/USDA_NAIP_DOQQ}} (agricultural monitoring data in the USA).

2) \textit{Amazon Web Services (AWS)}\footnote{\url{https://aws.amazon.com}} is a powerful platform for geospatial computing that offers a range of services for geospatial data storage, processing, and analysis. AWS hosts several representative geospatial datasets, including Digital Earth Africa\footnote{\url{https://www.digitalearthafrica.org}} (Landsat and Sentinel products over Africa), NOAA Emergency Response Imagery\footnote{\url{https://oceanservice.noaa.gov/hazards/emergency-response-imagery.html}} (LiDAR and hyperspectral data over the USA), and the datasets of the SpaceNet challenges (Tab.~\ref{tab:database}-\#134/264/298/299/300/304/306/312) \cite{van2018spacenet}. Sharing data on AWS allows anyone to analyze it and build services using a broad range of computing and data analytical products like Amazon EC2, which enables data users to spend more time on data analysis rather than data acquisition.

3) \textit{Microsoft Planetary Computer}\footnote{\url{https://planetarycomputer.microsoft.com}} provides access to a wide range of Earth observation data and powerful geospatial computing resources. The platform is specifically designed to support the development of innovative applications and solutions for addressing critical environmental challenges, such as climate change, biodiversity loss, and natural resource management. It offers cloud-based computing services that enable users to perform complex geospatial analyses efficiently. The Data Catalog in Planetary Computer provides access to petabytes of environmental monitoring data in a consistent and analysis-ready format. Some of the representative datasets available on the Planetary Computer Data Catalog include the HREA dataset\footnote{\url{https://planetarycomputer.microsoft.com/dataset/hrea}} (settlement-level measures of electricity access derived from satellite images) and the Microsoft Building Footprints dataset\footnote{\url{https://github.com/Microsoft/USBuildingFootprints}}.

4) \textit{Colab}\footnote{\url{https://colab.research.google.com}} provides a flexible geospatial computing platform that offers users access to free GPU and TPU (Tensor Processing Units) computing resources for analyzing geospatial data. It provides a free Jupyter notebook environment that allows users to write and run Python code for accessing and processing geospatial data from various platforms (e.g., GEE Data Catalog). Colab notebooks can be easily shared and collaborated on with others, which is particularly useful for geospatial analysis projects that involve multiple team members.

5) \textit{Kaggle}\footnote{\url{https://www.kaggle.com}} is an online platform that is widely used for data science and machine learning competitions. It provides a cloud-based computational environment that allows for reproducible and collaborative analysis in the field of geospatial computing. Kaggle also provides access to a variety of geospatial datasets, including satellite imagery, terrain data, and weather data, along with tools to analyze and visualize these datasets. Users can take advantage of the platform's free GPU and TPU computing resources to train machine learning models and undertake complex geospatial analysis tasks.
\section{Open Challenges and Future Directions}
\label{sec:open}

The previous sections give an overview of existing benchmarking datasets, presenting their main features and describing their main characteristics, eventually providing a broad yet detailed picture of the current state-of-the-art. This section discusses existing gaps, open challenges and potential future trends. 

\subsection{Where are we now?}
The most prominent issue that new dataset releases aim to fix is lack of diversity. 
Many of the earlier datasets contain samples that are extremely limited in their spatial and temporal distribution, to the extreme of consisting of a single, small image only. 
Not only are such datasets prone to leading to biased evaluation protocols, where information from the training set leaks into the test set via spatial correlation, they are usually also not sufficient to train models that are able to generalize to other geographic areas or time points (e.g. different seasons). 
More modern datasets aim to increase diversity regarding scene content (e.g. more different object instances), environmental factors (e.g. seasons, light conditions, etc.), or aspects regarding image acquisition / processing (e.g. different look angles, resolutions, etc.). 
This increase of diversity was so far always connected to an increase in dataset size by including more images and/or larger scenes or even other modalities. 
While an increase in the provided image data is often easily possible, it is usually not feasible to have the same increase in the reference data. 
This lead to large scale datasets where the reference data is much less carefully curated as for early datasets, often based on other existing sources (e.g. OpenStreetMap), and containing more label noise. 
While many machine learning methods can handle a certain extent of label noise during training, its effects on the evaluation are barely understood and often ignored.

\subsection{Current trends: Specificity and generality}
In this context, two main characteristics of a dataset for a given task will be the focus of the discussion: specificity and generality.

Earth Observation applications present numerous and diverse scenarios due to varying object types, semantic classes, sensor types and modalities, spatial, spectral, and temporal resolutions, and coverage (global, regional, or local). 

High specificity refers to datasets that are strongly tailored towards a specific sensor-platform-task combination, maybe even being limited to certain acquisition modes, geographic regions, or meteorological seasons.
These can hardly be used for anything beyond their original purpose.  
While different application domains, such as agriculture, urban mapping, military target detection, and bio-/geo-physical parameter extraction, do require different types of data, i.e., images, point measurements, tables, and metadata, the proliferation of different datasets specialized for every small task variation reduces the reusability of the datasets in a broader context and affects scientific transparency and reproducibility. 

A high specificity also contributes to cluttered nomenclature causing different datasets to appear different while actually sharing very similar content. 
For example, a high level of detail in class semantics and terminology makes it difficult to compare reference data of different datasets. 
A typical example is land cover classification, where similar classes may be aggregated into different subcategories depending on the application. 
As a result, models trained on different application-specific datasets may actually approximate very similar functional relationships between image data and target variables. 

Virtually all of the less recent and still most of the modern datasets aim for specificity. 
However, several of the more recent benchmarks follow another direction: generality, i.e. providing more sensor modalities than actually required plus large-scale, often noisy reference data for multiple tasks instead of small-scale and carefully curated annotations that only address a single task.

The contribution of such general datasets is manifold: 
First and foremost, the required number of (annotated) training samples for fully-supervised machine learning simply does not scale very well given the effort of data annotation and curation in remote sensing. 
Thus, such general, large-scale datasets introduce new factors that increase the relation to realistic application scenarios such as robustness to label noise (e.g. by leveraging existing semantic maps as reference data that are often outdated, misaligned, or of coarse resolution) and weakly-supervised learning (where the reference data has a lower level of detail than the actual target variable, e.g. training semantic segmentation networks with labels on image level). 
Large-scale datasets are the only option to realistically test the generalization capabilities of learning-based models, e.g. over different geographic regions, seasons, or other domain gaps. 
Furthermore, while multimodal datasets enable data-fusion and cross-modal approaches that leverage the different input sources, multi-task datasets allow exploiting the mutual overlap of related tasks regarding feature extraction and representation learning.
Last but not least, the idea of loosening the previously tight relationship between input data and target variable in datasets (up to the point where a dataset might not offer reference data for any target variable) is to provide data that can be leveraged to learn powerful representations that are useful for a large variety of downstream tasks (as in pre-training or self-supervised learning).

However, there is not yet a single "go-to" dataset that can be used for pre-training most newly developed models or for benchmarking specific tasks against state-of-the-art approaches. 
Collecting such a high-quality benchmark dataset that enables to pre-train models for as many downstream tasks as possible is of significant value for further pushing performance boundaries.

\noindent\fbox{%
    \parbox{\linewidth}{%
\textbf{The ideal pre-training dataset}\\
A dataset ideally suited for pre-training and/or self-supervised learning should adhere to as many of the following characteristics as possible:
\begin{itemize}
    \item Multiple platforms (vehicle, drone, airplane, satellite)
    \item Multiple sensors (Planet, SPOT, WorldView, Landsat, Sentinel~1/2, etc.)
    \item Several acquisition modalities (SAR, RGB, Hyperspectral, Multispectral, Thermal, LiDAR, Passive Microwave, etc.)
    \item Diverse acquisition geometries (viewing angles, e.g., off-nadir conditions, spatial and temporal baselines in multiview data, e.g., InSAR)
    \item Realistic distortion factors (cloud cover, dust, smog, fog, atmospheric influence, spatial misalignments and temporal changes in multi-view data, etc.)
    \item Well distributed geographical locations (spatial distribution within the dataset, climate zones, socio-economic and cultural factors, different topographies)
    \item Diverse land cover/use (urban, rural, forest, agricultural, water, etc.)
    \item Varying spatial resolution (0.1-1 m, 3-10 m, 10-30 m, 100-500 m, scale distribution)
    \item Well temporally distributed (seasonality, lighting condition, sun angle, nighttime imagery)
    \item Diverse set of reference data that are well aligned with the EO measurements (semantic information, change, geo-/bio-physical parameters, etc.)
\end{itemize}
}}

\begin{figure*}[ht]
    \centering
    \includegraphics[width=\linewidth]{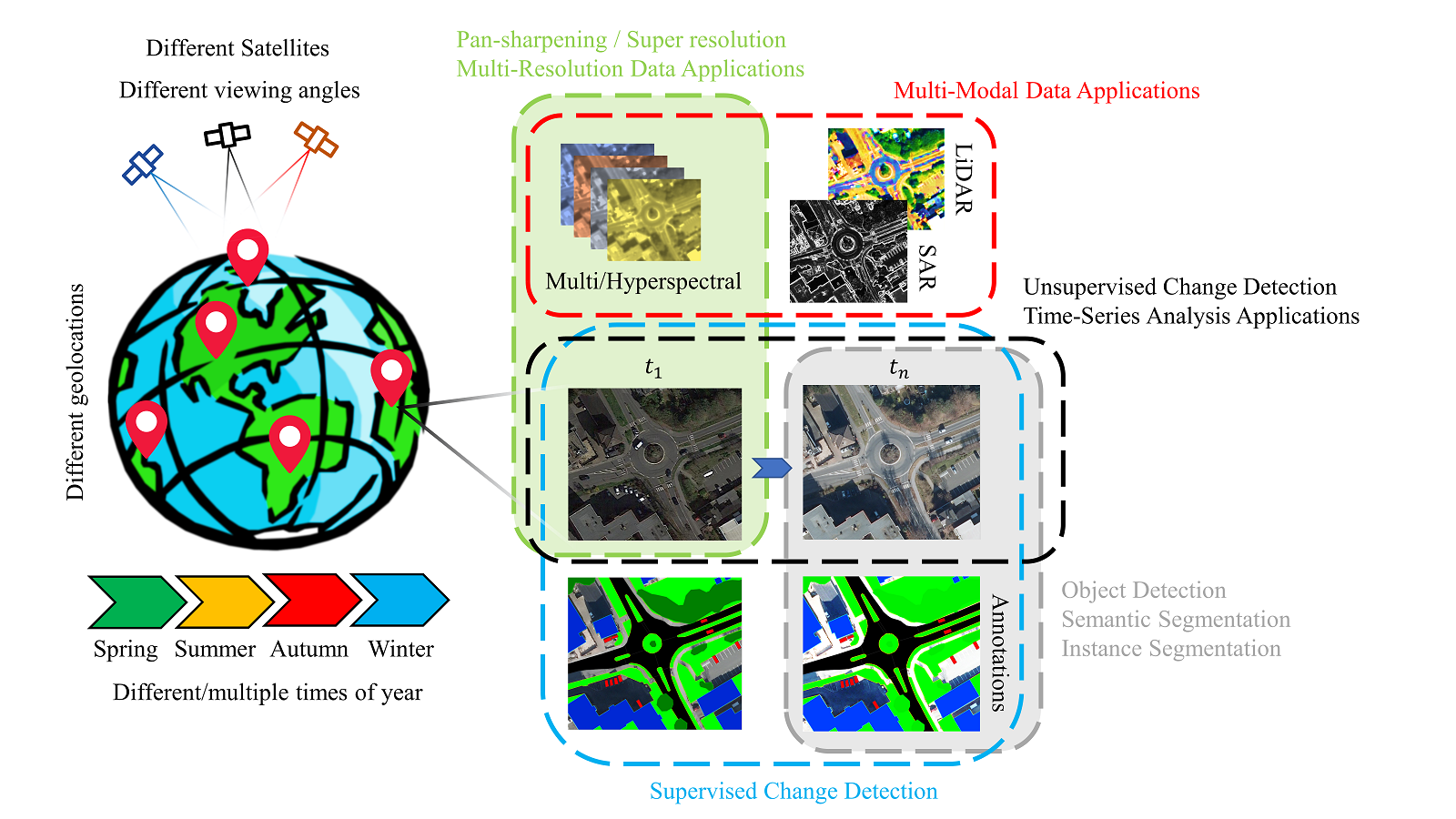}
    \caption{Illustration showing the authors' view of the paramount properties that an ideal benchmark dataset needs to satisfy, including the type of tasks, sensors, temporal constraints, and geolocalization.}
    \label{fig:future}
\end{figure*}

Fig.~\ref{fig:future} presents a schematic diagram of the properties of an ideal solution for a go-to EO benchmark dataset, covering diverse geo-locations, multiple modalities, different acquisition scenarios, and various applications. It is ideally acquired by different types of sensors and platforms with different viewing geometries to cover objects from different look-angles. The images are obtained from different electromagnetic spectrum bands, i.e., visible, infrared, thermal, and microwave, resulting in multi-/hyper-spectral, SAR, LiDAR, optical, thermal, and passive microwave measurements. The reference information or annotations are provided on a level that allows defining various tasks based on a single annotation. For example, an image with dense semantic annotations allows users to generate their desired object instance annotation files. 
Extending the dataset to multiple images of a scene with corresponding semantic labels enables not only semantic segmentation but also semantic change detection tasks. 

In summary, we foresee a certain duality in the future development of EO-related datasets: 
On the one hand, following the paradigm of data-centric machine learning \cite{Strickland2022}, i.e. moving against the current trend of creating performance gains merely by leveraging more training data but instead focusing on datasets tailored towards specific problems with well curated, high quality reference data (e.g. manually annotated or based on official sources). 
On the other hand, general datasets that cover as many in- and output modalities as possible to allow learning generic representations that are of value for a large number of possible downstream tasks.

\subsection{FAIR and ARD}
Additional to the content, scope, and purpose of datasets their organization will gain importance. 
With only a dozen public datasets available prior to 2015, it was feasible that each is provided with its own data format and meta-information, hosted on individual web pages, and downloaded by everyone who wanted to work with them. 
With the hundreds of datasets available today and many more published every year, this cannot be maintained. 
Concepts such as FAIR (Findability, Accessibility, Interoperability, and Reuse, see, e.g., \cite{Wilkinson2016}) have already been proposed years ago and are still of high relevance. 
Data catalogs such as EOD (see Section~\ref{sec:practical}) are a first step to structure datasets that are scattered among different data hosts.
ARD (Analysis Ready Data, see, e.g., \cite{Dwyer2018}), e.g. in the form of data cubes \cite{Giuliani2017}, and efforts to homogenize meta-information, e.g. in the form of datasheets \cite{Gebru2018}, will continue to evolve into new standardized data formats.
The trend of datasets growing in terms of size and volume (see Sec.~\ref{sec:evol}) as well as the need for global data products will soon put a stop on the current practice of downloading datasets and processing them locally. 
Working with data on distributed cloud services will create new requirements regarding data formats, but also lead to new standards and best practices for processing. 
Last but not least, the goal of any ML-centered dataset is to train an ML model. 
These models should be treated similar to the data they originated from, i.e. they should follow standardized data formats and FAIR principles. 
\section{Summary \& Conclusion}
This article discusses the relevance of machine-learning-oriented datasets in the technical disciplines of Earth observation and remote sensing. 
The analysis of the historical developments shows that the deep learning-boom has not only led to a rise in dataset numbers, but also large increase in size (as in spatial coverage and resolution but also with respect to multi-modal and multi-temporal imagery) and diversity of application tasks. This development has furthermore led to the implementation of dedicated software packages and metadatabases that help interested users to develop solutions for their applications. Eventually, we draw the conclusion that one of the critical challenges in dataset design for Earth observation tasks is the strong heterogeneity of possible sensors, data, and applications, which has led to a jungle of narrow-focused datasets. While one of the trends in deep learning certainly is the consideration of smaller, well-curated and task-specific datasets, another direction is the creation of a generic, nearly sensor- and task-agnostic database similar to the well-known ImageNet dataset used in Computer Vision. Such a generic dataset will especially be valuable in the pre-training of large high-capacity models with worldwide applicability. 
\clearpage
\begin{table*}[th!]
\begin{framed}
\section*{The Role of the IEEE-GRSS Data Fusion Contests}
The IEEE GRSS Data Fusion Contest (DFC) has been organized as an annual challenge since 2006 by the Image Analysis and Data Fusion Technical Committee (IADF TC) of the IEEE Geoscience and Remote Sensing Society (GRSS).
GRSS is an IEEE society with the goal of bringing together researchers and practitioners to monitor and understand Earth’s ecosystems and to identify potential risks. 
IADF is one of seven GRSS technical committees aiming at technical contributions within the scope of geospatial image analysis (e.g., machine learning, deep learning, image and signal processing, and big data) and data fusion (e.g., multi-sensor, multi-scale, and multi-temporal data integration). 
In general, the contest promotes the development of methods for extracting geospatial information from large-scale, multi-sensor, multimodal, and multi-temporal data. 
It focuses on challenging problem settings and establishes novel benchmark datasets for open problems in remote sensing image analysis (see Table~\ref{tab:dfc}).

Historically, the DFC developed from less formal activities related to the distribution of datasets between 2002 and 2006 by means of a collaboration between GRSS and the International Association of Pattern Recognition (IAPR). 
In 2006 the first DFC was organized by what was then called the Data Fusion Technical Committee of GRSS. 
It addressed the fusion of multispectral and panchromatic images, i.e., pan-sharpening, and provided one of the first public benchmark datasets for machine learning in remote sensing. 

Since 2006 various sensors have played a role in the DFC, 
including optical (SPOT \cite{dfc2009}, DEIMOS-2 \cite{dfc2016}, WorldView-3 \cite{dfc2019a}, aerial \cite{dfc2014,dfc2015a,dfc2015b,dfc2018,dfc2022,dfc2023}), 
multispectral (Landsat \cite{dfc2007,dfc2017,dfc2021msd,dfc2021dse}, Worldview-2 \cite{dfc2011,dfc2012}, Quickbird \cite{dfc2012}, Sentinel-2 \cite{dfc2017,dfc2020,dfc2021dse}, aerial \cite{dfc2021msd}), 
as well as hyperspectral \cite{dfc2008,dfc2013,dfc2018} images, 
SAR (ERS \cite{dfc2007,dfc2009}, TerraSAR-X \cite{dfc2012}, Sentinel-1 \cite{dfc2020,dfc2021dse}, Gaofeng \cite{dfc2023}) and 
LiDAR \cite{dfc2012,dfc2013,dfc2015a,dfc2015b,dfc2018,dfc2019b} data,
but also less common sources of Earth Observation data such as thermal images \cite{dfc2014}, 
digital elevation models \cite{dfc2022}, 
video-from-space (ISS \cite{dfc2016}),
night-time images (VIIRS \cite{dfc2021dse}), and 
Open Street Map \cite{dfc2017}.
Another meaningful change happened in 2017 when the DFC moved away from providing data over a single region only but instead allowed to train models over five cities worldwide (Berlin, Hong Kong, Paris, Rome, Sao Paulo) and testing on four other cities (Amsterdam, Chicago, Madrid, Xi’an.). 
This enabled creating models that generalize to new and unseen geographic areas instead of overfitting to a single location. 
This commendable trend was continued in 2020 by using SEN12MS (Tab.~\ref{tab:database}-\#87) \cite{sen12ms} as training data \cite{dfc2020}, providing data for the whole state of Maryland, USA, in 2021 (Tab.~\ref{tab:database}-\#1/291) \cite{dfc2021msd}, in total 19 different conurbations in France in 2022 (Tab.~\ref{tab:database}-\#241) \cite{dfc2022}, and data from 17 cities across six continents in 2023 \cite{dfc2023}.

The tasks addressed by the DFC over the years are dominated by 
semantic mapping \cite{dfc2007,dfc2008,dfc2013,dfc2014,dfc2015a,dfc2015b,dfc2017,dfc2018,dfc2019a,dfc2019b,dfc2021dse,dfc2023}, 
but also include pansharpening (Tab.~\ref{tab:database}-\#320) \cite{dfc2006}, change detection (in the context of floods) (Tab.~\ref{tab:database}-\#10) \cite{dfc2009}, and 3D reconstruction (Tab.~\ref{tab:database}-\#193) \cite{dfc2019a,dfc2023}, 
modern challenges such as weakly (Tab.~\ref{tab:database}-\#1/287/291) \cite{dfc2020,dfc2021msd} and semi-supervised learning (Tab.~\ref{tab:database}-\#241) \cite{dfc2022}, 
as well as open task contests \cite{dfc2011,dfc2012,dfc2014,dfc2016}, which allowed the participants to freely explore the potential of new and uncommon EO data. 

In 2006, seven teams from four different countries participated in the first DFC \cite{dfc2006} despite public contests being a new concept within the EO community. 
Being organized by an internationally well-connected society, providing exciting challenges, and establishing new benchmarks led to a quick increase in popularity. 
From seven teams in four countries in 2006, participation jumped quickly to 21 teams in 2008, 42 teams in 2014, and reached its peak with 148 teams (distributed over four different tracks, though) in 2019. 
The peak of the popularity was around 2012, which attracted more than 1000 registrations for the contest from nearly 80 different countries. 
Influenced by different factors, including the overwhelming success of datasets (and connected contests) in the Computer Vision community\footnote{Interestingly, the growing influence of Computer Vision on remote sensing, in particular due to deep learning, is also reflected by renaming the Data Fusion Technical Committee to the Image Analysis and Data Fusion (IADF) Technical Committee in 2014.}, an increasing amount of EO sensors with easier access to their data, as well as improved options for data hosting, the number of available benchmark datasets (that were not always but often connected to a contest) increased dramatically around 2015 (see also Figure~\ref{fig:evolution}). 
Since then, the participation in the DFC has been more or less constant (with a few positive outliers such as 2019), with around 40 teams from roughly 20 countries. 
Another commendable fact is that at the beginning the participants of the DFC were dominated by well-established scientists with solid experience in the respective fields. 
While this group still plays a vital role in more recent DFCs, a large number of participants (and winners!) are students. 
This shows that improved data availability helped to lower the starting threshold to analyze various Earth Observation data by standardizing data formats, easing access to theoretical knowledge, and open-sourcing software libraries and tools.

\vspace{10pt}
    \centering
\begin{tabular}{llll}
\toprule
  Year   & Data & Goal & Paper \\
  \cmidrule(r){1-1} \cmidrule(lr){2-2} \cmidrule(lr){3-3} \cmidrule(l){4-4}
    2023 & VHR optical and SAR satellite images & Classification, height estimation & \cite{dfc2023}\\
    2022 & VHR aerial optical images, DEM & Semi-supervised learning & \cite{dfc2022}\\
    2021 & Multi-temporal multispectral (aerial and Landsat 8) imagery & Weakly supervised learning & \cite{dfc2021msd}\\
         & Multispectral (Landsat-8, Sentinel-2), SAR (Sentinel-1), night-time (VIIRS) images & Semantic segmentation & \cite{dfc2021dse}\\
    2020 & Multispectral (Sentinel-2) and SAR (Sentinel-1) imagery & Weakly supervised learning & \cite{dfc2020}\\
    2019 & Optical (Worldview-3) images and LiDAR data & Semantic 3D reconstruction  & \cite{dfc2019a,dfc2019b}\\
    2018 & Multispectral LiDAR data, VHR optical and hyperspectral imagery & Semantic segmentation & \cite{dfc2018}\\
    2017 & Multispectral (Landsat, Sentinel-2) images and Open Street Map & Semantic segmentation & \cite{dfc2017}\\
    2016 & Very high temporal resolution imagery (DEIMOS-2) and video from space (ISS) & Open for creative ideas & \cite{dfc2016}\\
    2015 & Extremely high-resolution LiDAR and optical data & Semantic segmentation & \cite{dfc2015a,dfc2015b}\\
    2014 & Coarse resolution thermal/hyperspectral data and VHR color imagery & Semantic segmentation & \cite{dfc2014}\\
    2013 & Hyperspectral imagery and LiDAR-derived DSM & Semantic segmentation & \cite{dfc2013} \\
    2012 & VHR optical (QuickBird and WorldView-2), SAR (TerraSAR-X), and LiDAR data & Open for creative ideas & \cite{dfc2012} \\
    2011 & Multi-angular optical images (WorldView-2) & Open for creative ideas & \cite{dfc2011} \\
    2009-2010 & Multi-temporal optical (SPOT) and SAR (ERS) images & Change detection & \cite{dfc2009} \\
    2008 & VHR hyperspectral imagery & Semantic segmentation & \cite{dfc2008}\\
    2007 & Low-resolution SAR (ERS) and optical (Landsat) data & Semantic segmentation & \cite{dfc2007}\\
    2006 & Multispectral and panchromatic images & Pan-sharpening & \cite{dfc2006}\\
    \bottomrule
\end{tabular}
\caption{Overview of the IEEE GRSS Data Fusion Contests from 2006-23.}
\label{tab:dfc}
\end{framed}
\end{table*}




\bibliographystyle{IEEEtran}
\bibliography{ref}
%

%

%

\begin{IEEEbiographynophoto}{Michael Schmitt}
(michael.schmitt@unibw.de) received his Dipl.-Ing. (Univ.) degree in geodesy and geoinformation, his Dr.-Ing. degree in remote sensing, and his habilitation in data fusion from the Technical University of Munich (TUM), Germany, in 2009, 2014, and 2018, respectively. Since 2021, he has been a Full Professor for Earth Observation at the Department of Aerospace Engineering of the University of the Bundeswehr Munich in Neubiberg, Germany.
Before that, he was a professor for applied geodesy and remote sensing at the Munich University of Applied Sciences, Department of Geoinformatics. From 2015 to 2020, he was a senior researcher and deputy head at the Professorship for Signal Processing in Earth Observation at TUM; in 2019 he was additionally appointed as Adjunct Teaching Professor at the Department of Aerospace and Geodesy of TUM. In 2016, he was a guest scientist at the University of Massachusetts, Amherst. His research focuses on technical aspects of Earth observation, in particular image analysis and machine learning applied to the extraction of information from multi-sensor remote sensing observations. Among his core interests is remote sensing data fusion with a focus on SAR and optical data. He is a co-chair of the Working Group ``Active Microwave Remote Sensing'' of the International Society for Photogrammetry and Remote Sensing, and also of the Working Group ``Benchmarking'' of the IEEE-GRSS Image Analysis and Data Fusion Technical Committee. He frequently serves as a reviewer for a number of renowned international journals and conferences and has received several Best Reviewer awards. He is a Senior Member of the IEEE.
\end{IEEEbiographynophoto}

\begin{IEEEbiographynophoto}{Seyed Ali Ahmadi} received his B.Sc. degree in surveying engineering and M.Sc. degree in remote sensing from the Faculty of Geodesy and Geomatics, K. N. Toosi University of Technology, Tehran, Iran, in 2015 and 2017, respectively. He worked on image classification and segmentation techniques, machine learning algorithms, and LiDAR data processing. His thesis was focused on classifying hyperspectral and LiDAR datasets by combining spectral and spatial features in order to increase the classification accuracy. He is currently pursuing his Ph.D. thesis on building damage assessment with the K. N. Toosi University of Technology. His research interests include machine learning, deep learning, geospatial data analysis, image processing, and computer vision techniques for remote sensing and earth observation applications. He is a co-chair of Working Group ``Benchmarking'' of the IEEE-GRSS Image Analysis and Data Fusion Technical Committee; and frequently serves as a reviewer for a number of international journals and has received a Best Reviewer Award in 2018.
\end{IEEEbiographynophoto}

\begin{IEEEbiographynophoto}{Yonghao Xu} received the B.S. degree and the Ph.D. degree in Photogrammetry and Remote Sensing from Wuhan University, Wuhan, China, in 2016 and 2021, respectively. He is currently a post-doctoral researcher at the Institute of Advanced Research in Artificial Intelligence (IARAI), Austria. His research interests include remote sensing, computer vision, and machine learning.
\end{IEEEbiographynophoto}

\begin{IEEEbiographynophoto}
{Gulsen Taskin}
(SM’22-M'10) received the B.S. degree in Geomatics Engineering, the M.S. degree in Computational Science and Engineering (CSE), and the Ph.D. degree in CSE, from Istanbul Technical University (ITU), Turkey, in 2001, 2003, and 2011, respectively.
She is currently an Associate Professor at the Institute of Disaster Management at Istanbul Technical University. She was a visiting scholar at the School of Electrical and Computer Engineering and the School of Civil Engineering at Purdue University from 2008 to 2009 and 2016 to 2017. 
 She is a reviewer for Photogrammetric Engineering and Remote Sensing, IEEE Transactions on Geoscience and Remote Sensing,  IEEE Journal of Selected Topics in Applied Earth Observations and Remote Sensing, IEEE Geoscience and Remote Sensing Letters, and IEEE Transactions on Image Processing. Her current interests include machine learning approaches in hyperspectral image analysis, dimensionality reduction,  explainable AI, and sensitivity analysis. 
 \end{IEEEbiographynophoto}
 
 \begin{IEEEbiographynophoto}{Ujjwal Verma} received his Ph.D. from Télécom ParisTech, University of Paris-Saclay, Paris, France, in Image Analysis and his M.S. (Research) from IMT Atlantique (France) in Signal and Image Processing. Dr. Verma is currently an Associate Professor and Head of the Department of Electronics and Communication Engineering at Manipal Institute of Technology, Bengaluru, India. His research interests include Computer Vision and Machine Learning, focusing on variational methods in image segmentation, deep learning methods for scene understanding, and semantic segmentation of aerial images. He is a recipient of the "ISCA Young Scientist Award 2017-18" by the Indian Science Congress Association (ISCA),  a professional body under the Department of Science and Technology, Government of India. Dr. Verma is also a recipient of the "Young Professional Volunteer Award 2020" by IEEE Mangalore Sub-Section in recognition of his outstanding contribution to IEEE activities. Dr. Verma is the Co-Lead for the Working Group on Machine/Deep Learning for Image Analysis (WG-MIA) of the Image Analysis and Data Fusion Technical Committee (IADF TC) of the IEEE Geoscience and Remote Sensing Society. He is Guest Editor for Special Stream in IEEE Geoscience and Remote Sensing Letters and a reviewer for several journals (IEEE Transactions on Image Processing, IEEE Transactions on Geoscience and Remote Sensing, IEEE Geoscience and Remote Sensing Letters). He is also a Sectional Recorder for the ICT Section of the Indian Science Congress Association for 2020-24. 
 \end{IEEEbiographynophoto}

\begin{IEEEbiographynophoto}{Francescopaolo Sica}
received the Laurea (M.S.) degree (summa cum laude) in Telecommunications Engineering and the Dr. Ing. (Ph.D.) degree in Information Engineering from the University of Naples Federico II, Italy, in 2012 and 2016, respectively. Since 2022, he is Deputy Head of the Earth Observation Laboratory at the Department of Aerospace Engineering of the University of the Bundeswehr Munich, Germany. Between 2016 and 2022, he was a researcher at the German Aerospace Center (DLR). His research interests cover a wide range of activities related to Synthetic Aperture Radar (SAR) technology, from mission design to SAR signal and image processing to end-user applications. Dr Sica received a Living Planet Postdoc Fellowship from the European Space Agency (ESA) for the HI-FIVE (High-Resolution Forest Coverage with InSAR \& Deforestation Surveillance) project. He is co-chair of the Benchmarking Working Group of the IEEE-GRSS Image Analysis and Data Fusion Technical Committee (IADF) and is a regular reviewer for international journals and conferences.
\end{IEEEbiographynophoto}

\begin{IEEEbiographynophoto}{Ronny H\"{a}nsch} 
received the Diploma in computer science and the Ph.D. degree from the TU Berlin, Berlin, Germany, in 2007 and 2014, respectively. He is a scientist at the Microwave and Radar Institute of the German Aerospace Center (DLR) where he leads the Machine Learning Team in the Signal Processing Group of the SAR Technology Department. He continues to lecture at the TU Berlin in the Computer Vision and Remote Sensing Group. His research interest is computer vision and machine learning with a focus on remote sensing (in particular SAR processing and analysis). He serves as chair of the GRSS Image Analysis and Data Fusion (IADF) technical committee, co-chair of the ISPRS working group on Image Orientation and Fusion, editor of the GRSS eNewsletter, GRSS membership chair, associate editor of the Geoscience and Remote Sensing Letters and the ISPRS Journal of Photogrammetry and Remote Sensing, organizer of the CVPR Workshops EarthVision (2017-2023), Photogrammetric Computer Vision (2019,2023), the ML4RS Workshop at ICLR (2023), and the IGARSS Tutorial on Machine Learning in Remote Sensing (2017-2023). He has extensive experience in organizing remote sensing community competitions (e.g. the DFC 2018-2023), serves as the GRSS representative within SpaceNet, and was the technical lead of the SpaceNet 8 Challenge. 
\end{IEEEbiographynophoto}




\end{document}